\documentclass[10pt,twocolumn,letterpaper]{article}

\usepackage{wacv}              

\usepackage{enumitem}
\usepackage{graphicx}
\usepackage{amssymb}
\usepackage{booktabs}
\usepackage{tabularx}
\usepackage{wrapfig}
\usepackage{caption}
\usepackage{array}
\usepackage{amsthm}
\usepackage{graphics}
\usepackage{adjustbox}
\usepackage{makecell}
\usepackage{multirow}
\usepackage{soul}
\usepackage{svg}
\usepackage{tikz}
\usepackage{amsmath}
\usepackage{xspace}
\usepackage{siunitx}
\usepackage{mathtools}
\usepackage{algorithm}
\usepackage{algorithmicx}
\usepackage[noend]{algpseudocode}
\usepackage{subcaption}
\usepackage{xparse}
\usepackage{dsfont}
\usepackage{colortbl}
\usepackage{float}
\usepackage{bm}
\usepackage{pifont}
\usepackage{dashbox}
\usepackage{sourcecodepro}
\usepackage{longtable}
\usepackage[most]{tcolorbox}
\usepackage{url}
\usepackage{amsfonts}
\usepackage{nicefrac}
\usepackage{microtype}
\usepackage{xcolor}
\usepackage{changepage}
\usepackage{rotating}
\usepackage{wrapfig}

\usepackage[utf8]{inputenc} 
\usepackage[T1]{fontenc}    

\let\olduparrow\uparrow
\renewcommand{\uparrow}[1][1pt]{%
  \mathrel{\raisebox{#1}{$\olduparrow$}}%
}

\algnewcommand\algorithmicforeach{\textbf{for each}}
\algdef{S}[FOR]{ForEach}[1]{\algorithmicforeach\ #1\ \algorithmicdo}

\definecolor{SpringGreen}{HTML}{ebf3e9}
\definecolor{OliveGreen}{HTML}{bdd7af}
\definecolor{darkbrown}{HTML}{7c562a}

\usepackage{graphicx}
\usepackage{amsmath}
\usepackage{amssymb}
\usepackage{booktabs}

\usepackage[pagebackref,breaklinks,colorlinks]{hyperref}

\usepackage[capitalize]{cleveref}
\crefname{section}{Sec.}{Secs.}
\Crefname{section}{Section}{Sections}
\Crefname{table}{Table}{Tables}
\crefname{table}{Tab.}{Tabs.}

\def\wacvPaperID{2240} 
\def\confName{WACV}
\def\confYear{2026}

\begin{document}

\newcommand{\ours}{microCLIP\xspace}

\title{\includegraphics[height=1em]{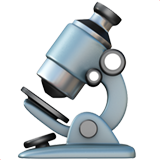} microCLIP: Unsupervised CLIP Adaptation via Coarse-Fine Token Fusion for Fine-Grained Image Classification}

\author{
Sathira Silva$^{1}$ \qquad
Eman Ali$^{1,2}$ \qquad
Chetan Arora$^{3}$ \qquad
Muhammad Haris Khan$^{1}$ \\
$^{1}$Mohamed Bin Zayed University of Artificial Intelligence \qquad
$^{2}$Alexandria University \qquad
$^{3}$IIT Delhi 
}

\maketitle

\begin{abstract}


Unsupervised adaptation of CLIP-based vision-language models (VLMs) for fine-grained image classification requires sensitivity to microscopic local cues. While CLIP exhibits strong zero-shot transfer, its reliance on coarse global features restricts its performance on fine-grained classification tasks. Prior efforts inject fine-grained knowledge by aligning large language model (LLM) descriptions with CLIP's $\texttt{[CLS]}$ token; however, this approach overlooks spatial precision. We propose $\textbf{microCLIP}$, a self-training framework that jointly refines CLIP's visual and textual representations using fine-grained cues. At its core is Saliency-Oriented Attention Pooling (SOAP) within a lightweight TokenFusion module, which builds a saliency-guided $\texttt{[FG]}$ token from patch embeddings and fuses it with the global $\texttt{[CLS]}$ token for coarse-fine alignment. To stabilize adaptation, we introduce a two-headed LLM-derived classifier: a frozen classifier that, via multi-view alignment, provides a stable text-based prior for pseudo-labeling, and a learnable classifier that is initialized from LLM descriptions and fine-tuned with TokenFusion. We further develop Dynamic Knowledge Aggregation, which convexly combines fixed LLM/CLIP priors with TokenFusion's evolving logits to iteratively refine pseudo-labels. Together, these components uncover latent fine-grained signals in CLIP, yielding a consistent $\mathbf{2.90\%}$ average accuracy gain across 13 fine-grained benchmarks while requiring only light adaptation. Our code is available at \href{https://github.com/sathiiii/microCLIP}{https://github.com/sathiiii/microCLIP}.

\end{abstract}    
\section{Introduction}

\textbf{CLIP's Global Objective:} Recent advances in foundation vision-language models (VLMs)~\cite{align, ALBEF, BLIP, FLAVA, li2022supervision, xu2023demystifying} have reshaped zero-shot learning, with CLIP~\cite{clip} emerging as a straightforward yet powerful approach. CLIP is pretrained with a contrastive objective on image-caption pairs by aligning global image representations, typically the \texttt{[CLS]} token, with sentence-level text embeddings in a shared embedding space. This alignment strategy enables CLIP to capture high-level, coarse-grained semantics, supporting strong generalization to classification tasks in domains different from its pretraining data. As a result, CLIP demonstrates impressive zero-shot transfer in training-free methods~\cite{zhang2022tip, CuPL, wca}, and can be further adapted to downstream tasks using few-shot~\cite{coop, khattakMaPLe, lafon2024gallop, lin2024tagclip} or unlabeled samples in an \textit{Unsupervised Adaptation} (UA) setting~\cite{upl, pouf, lafter, reclip, DPA}.

\begin{figure}[t]
    \centering
    \includegraphics[width=\linewidth]{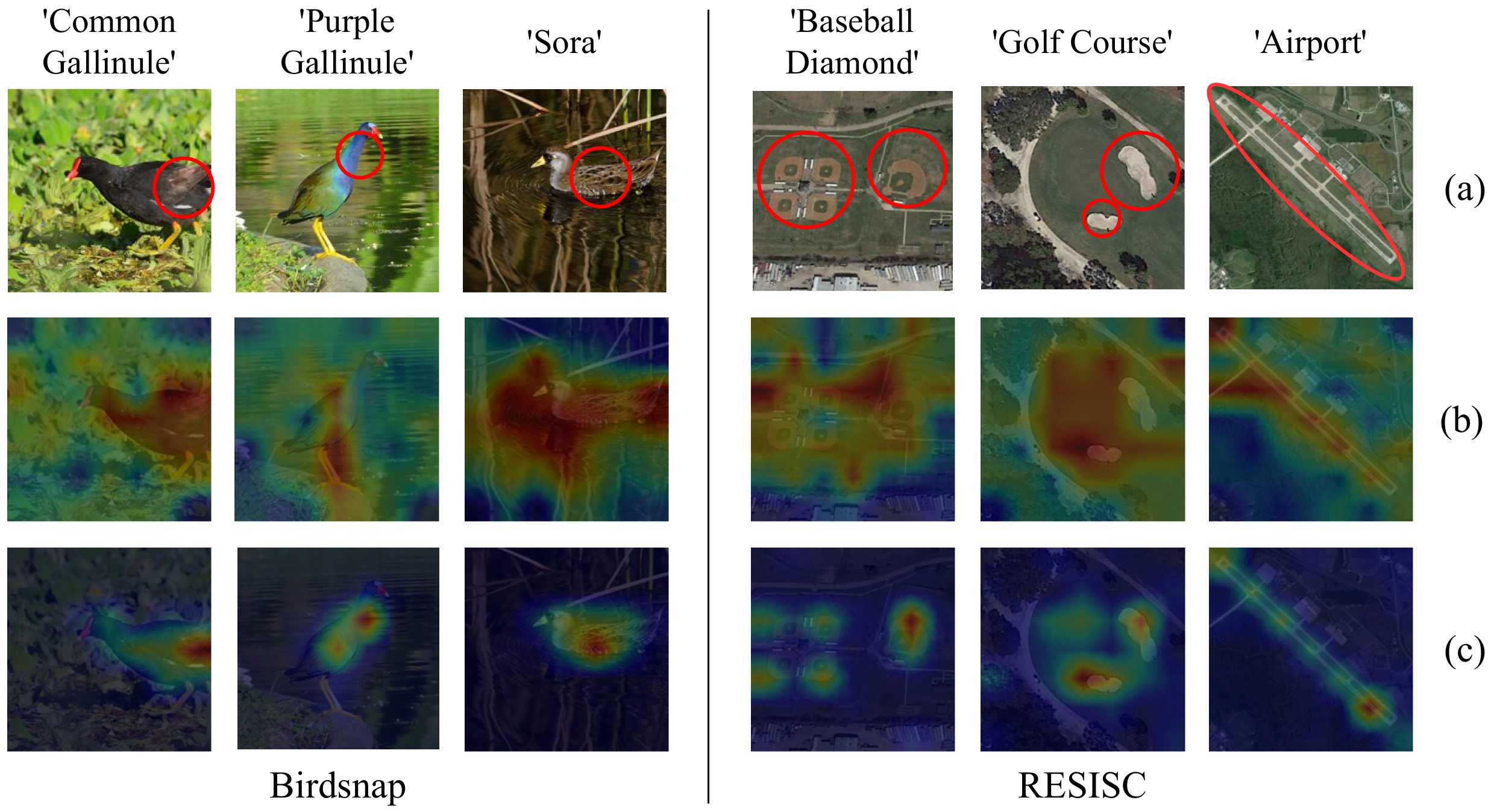}
    \caption{Attention maps on two fine-grained datasets: Birdsnap and RESISC45. Row (a): input images; (b): global attention from DPA~\cite{DPA}; (c): local attention from \textbf{microCLIP} (ours). By guiding the \texttt{[FG]} token with SOAP queries, microCLIP focuses on semantically critical regions, yielding sharper, more discriminative attention. Red circles highlight referenced regions in the text.}
    \label{fig:teaser_motivation}
\end{figure}

\noindent\textbf{Gaps in UA Literature:} Fine-grained image classification~\cite{krause20133d,nilsback2008automated,parkhi2012cats,birdsnap} aims to differentiate between closely related categories by focusing on subtle, localized visual details. Despite CLIP's strong shared embedding space, the coarse granularity of CLIP's visual representation prevents it from capturing fine, local discriminative features essential for fine-grained classification. Such limitations become pronounced in realistic scenarios where no labeled data exist and the model must generate pseudo-labels on its own. Early improvements in zero-shot performance involved domain-specific prompt ensembles~\cite{clip} or LLM-generated text descriptions~\cite{CuPL,encoder,fan2024improving} to better align with categories.
Previous UA methods, such as LaFTer~\cite{lafter}, incorporate fine-grained knowledge priors from LLM-generated descriptions, whereas others, like DPA~\cite{DPA}, build coarse-grained visual priors by caching image prototypes from unlabeled data.
We argue that these methods are limited by their reliance on CLIP’s pretrained \texttt{[CLS]} token, which aligns poorly with fine-grained textual descriptions (see Table 1 in the supplementary). This coarse representation often misses \textit{local semantics}, spatial cues crucial for distinguishing subtle differences. As shown in~\cref{fig:teaser_motivation} (middle row), DPA’s attention maps frequently highlight irrelevant regions, resulting in suboptimal performance on fine-grained tasks~\cite{detailclip, kim2025cosmos, xiao2024flair}. To address the limitations of the coarse-grained \texttt{[CLS]} token, WCA~\cite{wca}, a training-free method, aligns LLM-generated descriptions with multiple random local image views iteratively. We find that using multi-view representations as weak augmentation improves offline pseudo-labeling. Notably, we reduce the number of local views (by roughly $8\times$) without sacrificing the pseudo-label quality. 
\noindent\textbf{Our Contributions:} We show that relying solely on fine-grained cues from text for unsupervised adaptation is inherently limited and provide empirical evidence for this. While the coarse \texttt{[CLS]} token may miss local details, it preserves valuable global knowledge from CLIP pretraining. Rather than discarding it, we treat \texttt{[CLS]} as a strong global prior, augmenting it with fine-grained cues from patch tokens. Motivated by the limitations mentioned above and inspired by recent attention-pooling methods~\cite{xiao2024flair, zheng2024dreamlip}, we propose \textbf{\ours}. This self-training framework jointly refines CLIP’s textual and visual representations, injecting LLM-derived textual priors and enhancing visual features with localized cues. To our knowledge, this is the first UA method to coordinate the fine-tuning of both modalities with fine-grained information. To summarize, we make the following contributions:

\begin{itemize}
    \item A novel \emph{Saliency-Oriented Attention Pooling (SOAP)} mechanism within our lightweight \emph{TokenFusion} module, which builds a saliency query on CLIP patch tokens to pool a compact \texttt{[FG]} token; TokenFusion then fuses \texttt{[FG]} with CLIP’s global \texttt{[CLS]} for coarse–fine alignment.
    \item A \emph{two-headed LLM-derived classifier}: a frozen LLM-derived classifier $W_{LLM}$ that, via multi-view alignment, provides a stable text-based prior for pseudo-label generation, and a learnable classifier $W^*_{LLM}$ (initialized from LLM description) that is fine-tuned with TokenFusion.
    \item We propose \emph{Dynamic Knowledge Aggregation}, an iterative pseudo-labeling scheme that convexly combines fixed CLIP/LLM priors obtained through multi-view alignment with TokenFusion's evolving logits, enabling stable yet adaptive self-training for fine-grained distinctions.
\end{itemize}

We empirically show these components reveal CLIP’s latent fine-grained signals, producing an average gain of \(\mathbf{+2.90\%}\) across 13 fine-grained datasets with only lightweight adaptation. Our saliency-based localized attention consistently highlights class-defining \emph{local semantics} (see \cref{fig:teaser_motivation}, bottom): e.g., the reddish-brown body of the `Common Gallinule', the purple neck of the `Purple Gallinule', and the dark feathers of the `Sora' in Birdsnap~\cite{birdsnap}; and the infield layout of `Baseball Diamond', sandy areas of `Golf Courses', and runways of `Airport' in RESISC~\cite{Cheng2017RemoteSI}.

\section{Related Works}

\begin{figure*}[t]
    \centering
    \includegraphics[width=\linewidth, trim=0 0 50 0, clip]{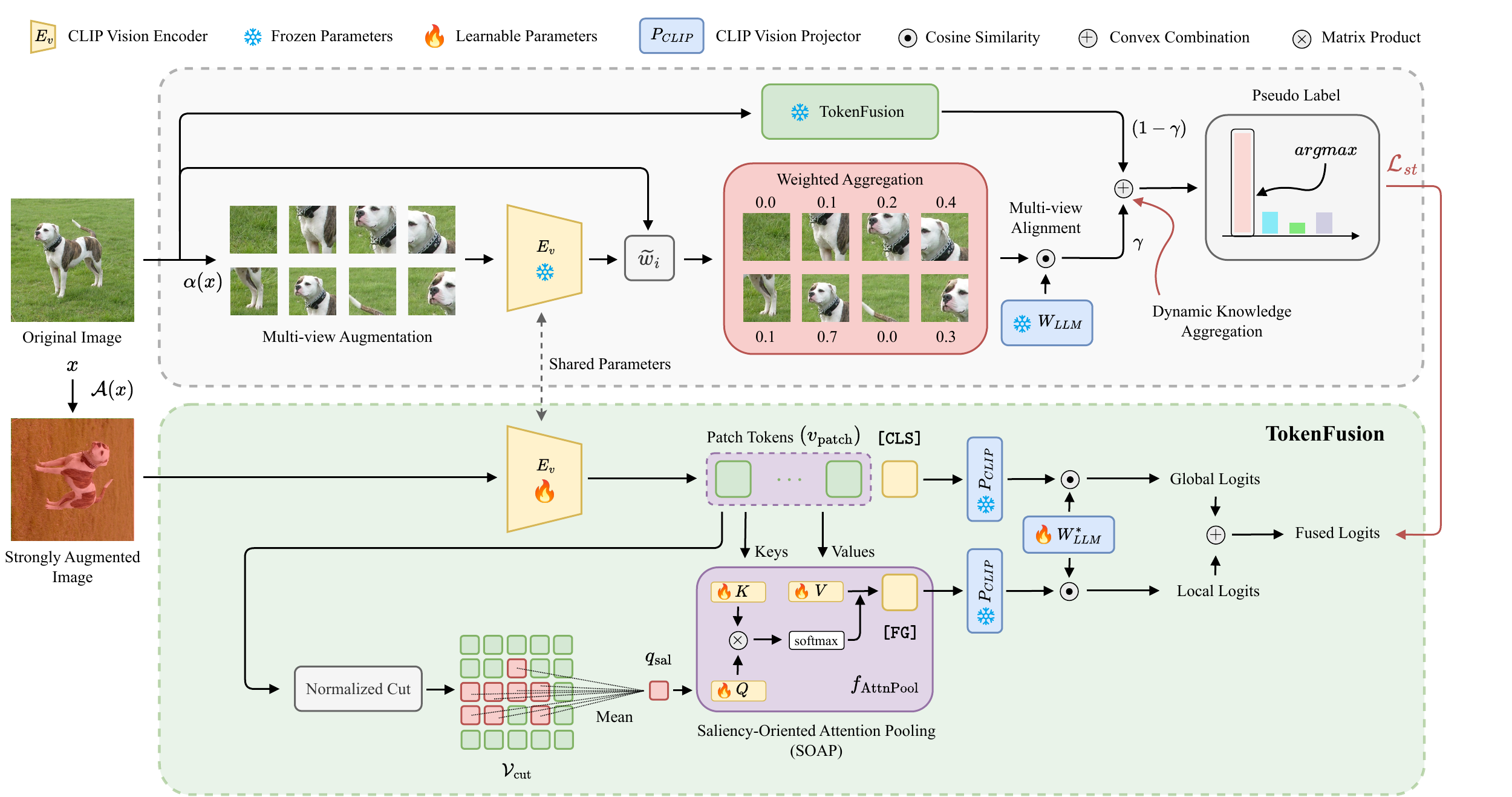}
    \caption{
        \textbf{Overall architecture of \ours.} The top shows our 
        \tcbox[colback=gray!10, colframe=gray!50, boxrule=0.3pt, arc=0.8mm, left=0.5pt, right=0.5pt, top=0pt, bottom=0pt, on line]{pseudo-labeling} 
        pipeline, where fixed knowledge from CLIP via the alignment between multi-view augmented representations and fine-grained LLM-generated descriptions is combined with dynamic knowledge learned in TokenFusion. The bottom illustrates our 
        \tcbox[colback=SpringGreen, colframe=OliveGreen, boxrule=0.3pt, arc=0.8mm, left=0.5pt, right=0.5pt, top=0pt, bottom=0pt, on line]{TokenFusion} module.
    }
    \label{fig:architecture}
\end{figure*}

\textbf{Unsupervised Adaptation of CLIP:} 
CLIP~\cite{clip} employs contrastive learning to align images and text in a shared latent space, enabling robust zero-shot learning. However, unsupervised adaptation (UA) of CLIP for fine-grained downstream tasks remains challenging. Existing work like UPL~\cite{upl} utilizes top-K pseudo-labeling for unsupervised prompt learning, while POUF~\cite{pouf} aligns prototypes with target data using transport-based distribution alignment. LaFTer~\cite{lafter} fine-tunes a visual prompt with LLM-generated texts and unlabeled images. ReCLIP~\cite{reclip} tackles visual-text misalignment via a projection space and fine-tunes both encoders simultaneously while investing in costly label propagation for pseudo-labeling. DPA~\cite{DPA} improves pseudo-labeling accuracy by aligning visual and textual prototypes to reduce noise, using a prototypical classifier initialized with handcrafted, prompt-ensembled textual prototypes that are then fine-tuned. Despite these efforts, fine-grained unsupervised adaptation of CLIP remains an unresolved challenge.  
In this work, we adapt CLIP using saliency-guided attention pooling with the \texttt{[CLS]} token to align visual cues with fine-grained textual cues provided by an LLM-derived prototypical classifier.


\noindent\textbf{Multi-view Representations:} DINO-MC~\cite{wanyan2023dinomc} extends global–local contrastive learning using multi-scale crops, enriching CLIP representations with fine-grained spatial context. VCR~\cite{lu2024rethinkingvisualcontentrefinement} selects confident multi-scale crops to construct robust features that better align with textual descriptions. WCA~\cite{wca} introduces a Visual-Text Cross-Alignment strategy that randomly samples a large number of image crops and aggregates predictions through similarity-weighted averaging. While effective, WCA is computationally expensive. Inspired by WCA’s use of diverse views to enhance alignment, we adopt a more efficient alternative: treating a small set of multiple local views as a weak augmentation and directly aligning them with a set of LLM-derived class prototypes.

\noindent\textbf{Extraction of Salient Regions:}
Unsupervised salient object detection aims to identify prominent regions without annotations. Earlier approaches~\cite{borji2019salient, perazzi2012saliency, liu2010learning, yang2013saliency, zhu2014saliency} relied on handcrafted features like contrast and boundary priors but struggled in complex scenes. Recent self-supervised methods such as SelfMask~\cite{Shin2022UnsupervisedSO} and FOUND~\cite{simeoni2023unsupervised} exploit deep features but offer only binary region separation. Instead, we adopt the graph-theoretic Normalized Cut (NCut) framework~\cite{NCut}, following TokenCut~\cite{TokenCut}, to segment informative patch tokens by capturing instance-level saliency in the feature similarity graph.

\noindent\textbf{Utilization of Patch Tokens in Fine-Grained Tasks:}
Patch tokens have traditionally been employed in segmentation-focused research. Studies such as~\cite{zhou2022extract, wang2024sclip, ranasinghe2023perceptual} demonstrate that CLIP effectively captures object appearance but faces challenges in spatial localization due to its global attention mechanism. Works like MaskCLIP~\cite{zhou2022extract} and SCLIP~\cite{wang2024sclip} enhance token-level spatial cues by modifying attention pooling or strengthening correlations. Our work introduces a locally aggregated fine-grained token, \texttt{[FG]}, repurposing patch tokens for fine-grained classification via a saliency-guided aggregation, using only CLIP’s pretrained features.

\section{Methodology}

\subsection{Preliminaries}

Our study addresses the challenge of adapting the CLIP model for fine-grained image classification without requiring labeled data. We leverage the pretrained CLIP, which comprises a visual encoder $E_v$ and a textual encoder $E_t$. 
In our experimental framework, we consider a dataset $ \mathcal{D}_t = \{ \mathcal{X}_t \} $, which consists of unlabeled images $x$, where each $ x \in \mathcal{X}_t $. Additionally, we assume the availability of unique class names $ y \in \mathcal{Y} $ for $\mathcal{D}_t$. 

\subsection{Overall Architecture}

As illustrated in \cref{fig:architecture}, \ours comprises two key components: (1) the \textit{TokenFusion} module based on \textit{Saliency-Oriented Attention Pooling}, and (2) an iteratively improving pseudo-labeler based on \textit{Dynamic Knowledge Aggregation}. To induce CLIP to reveal fine-grained cues, we initialize a two-headed LLM-derived classifier: a frozen classifier \(W_{LLM}\) used as a stable multi-view prior for pseudo-labeling and a learnable classifier \(W_{LLM}^*\) (initialized from the same descriptions) that is fine-tuned with TokenFusion. The text encoder \(E_t\) is used only for initialization and is discarded thereafter. In the following sections, we provide a detailed explanation of each component.

\subsection{TokenFusion Module}
\textbf{Saliency-Oriented Attention Pooling (SOAP)
for CLIP:} 
Prior work on fine-grained classification with CLIP often overlooks patch tokens in the vision encoder, focusing solely on the \texttt{[CLS]} token, as CLIP explicitly optimizes the \texttt{[CLS]} token while patch tokens contribute implicitly via the attention mechanism.
Recent VLM pretraining approaches~\cite{zheng2024dreamlip, xiao2024flair} aim to build fine-grained cross-modal representations using patch tokens to enable region-specific understanding. However, these methods depend on large-scale image-text corpora. TagCLIP~\cite{lin2024tagclip}, inspired by GradCAM~\cite{GradCAM}, highlights the importance of patch tokens, showing that the penultimate layer of the CLIP image encoder retains spatial details absent in the final layer. A naive way to aggregate these would be to apply average pooling over the tokens; however, this introduces noise into the intended fine-grained representation and leads to degraded performance in UA (see ablations in \cref{tab:attention_pooling}).
FLAIR~\cite{xiao2024flair} uses multi-head attention pooling with fine-grained captions as queries to pretrain cross-modal attention. In contrast, \ours introduces a novel \textit{Saliency-Oriented Attention Pooling} (SOAP) mechanism that uses the Normalized Cut (NCut) algorithm~\citep{NCut} to filter out noisy tokens and isolate salient ones. These tokens, already enriched with positional encoding, are averaged to form a query that guides attention toward the most informative CLIP patch embeddings, improving spatial awareness and fine-grained representation.

Formally, the CLIP image encoder $E_v$ processes an input image $x$ and produces $n$ local patch tokens along with a global token, \texttt{[CLS]}, represented as $E_v(x) = [x_{\text{patch}}, v^{\texttt{CLS}}]$, where $x_{\text{patch}} \in \mathbb{R}^{n \times d}$ denotes the patch tokens and $v^{\texttt{CLS}} \in \mathbb{R}^{d}$ is the global representation of dimension $d$.
To retain spatial information, we derive $v_{\text{patch}}$ by bypassing attention~\cite{lin2024tagclip}, as in~\cref{eq:bypass_attn}. The resulting tokens are then forwarded through the remaining layers, as given in \cref{eq:v_patch}, rather than using the patch tokens from the final output of $E_v$. 
\begin{align}
    \label{eq:bypass_attn}
    \tilde{v}_{\text{patch}} &= x_{\text{patch}}^{L-1} + x_{\text{patch}}^{L-1}\widetilde{W}_V^L\\
    \label{eq:v_patch}
    v_{\text{patch}} &= \tilde{v}_{\text{patch}} + \text{MLP}(\tilde{v}_{\text{patch}})
\end{align}
Here, $L$ denotes the number of layers in $E_v$, $\widetilde{W}_V^L$ is the value projection matrix at the final layer, and MLP refers to the multilayer perceptron module used in that layer. We treat the patch tokens as nodes in a fully connected graph, where edges represent pairwise token similarities. We then apply the NCut algorithm~\cite{NCut} to select a subset of tokens corresponding to the image’s most salient regions, denoted by $\mathcal{V}_{\text{cut}}$, as shown in~\cref{eq:v_cut}. 
Implementation details for NCut are provided in the supplementary materials (Appendix D). Since $v_{\text{patch}}$ already encodes rich spatial information via positional embeddings (introduced before the step expressed in \cref{eq:bypass_attn}), we simply average the tokens in $\mathcal{V}_{\text{cut}}$ to obtain a saliency-aware query vector, $q_{\text{sal}}$, as described in \cref{eq:query}.
\begin{align}
\label{eq:v_cut}
\mathcal{V}_{\text{cut}} &= \text{NCut}(v_{\text{patch}})\\
\label{eq:query}
q_{\text{sal}} &= \frac{1}{|\mathcal{V}_{\text{cut}}|} \sum_{\forall v \in \mathcal{V}_{\text{cut}}} v
\end{align}
The query $q_{\text{sal}}$ guides the attention pooling module $f_{\text{AttnPool}}$ to produce the fine-grained \texttt{[FG]} token $v^{\texttt{FG}} \in \mathbb{R}^d$:
\begin{equation}
\begin{aligned}
v^{\texttt{FG}} &= f_{\text{AttnPool}}(q_{\text{sal}}, v_{\text{patch}})\\
&= \text{softmax} \left( \frac{q_{\text{sal}} W_{\text{Q}} (v_{\text{patch}} W_{\text{K}})^\top}{\sqrt{d}} \right) v_{\text{patch}} W_{\text{V}}
\end{aligned}
\label{eq:atten}
\end{equation}
In \cref{eq:atten}, $W_Q$, $W_K$, and $W_V$ denote the query, key, and value projection matrices, respectively, and $d$ is the token embedding dimension. We implement $f_{\text{AttnPool}}$ as a single-head attention layer, as $q_{\text{sal}}$ already encodes spatial and contextual cues inherited from pretrained CLIP. This eliminates the need for multi-head attention, commonly used to model diverse representation subspaces, and reduces computational overhead. We append an empty token to $v_{\text{patch}}$, enabling $q_{\text{sal}}$ to attend to it in cases where $q_{\text{sal}}$ and $v_{\text{patch}}$ may not be semantically well aligned~\cite{xiao2024flair}.

\noindent\textbf{TokenFusion for Granularity-Enhanced Representation:} Our TokenFusion module leverages the $v^{\texttt{FG}}$ token, generated through SOAP, to capture region-specific visual details critical for fine-grained classification. Note that since we operate in the same visual embedding space of CLIP during the creation of $v^{\texttt{FG}}$, this enables us to treat it similarly to the global $v^{\texttt{CLS}}$ token and, therefore, use CLIP's learned projection, $P_{\text{CLIP}}$, to project $v^{\texttt{FG}}$ from the vision space to the shared embedding space. 

Unlike traditional approaches that rely solely on coarse-grained global visual embeddings to align with textual embeddings~\cite{lafter, reclip, DPA}, our method posits that fine-grained classification benefits from a combination of both local and global visual features, and thus computes predictions by fusing the two. To compute local logits, we utilize the $v^{\texttt{FG}}$ token and project it onto the shared embedding space using $P_{\text{CLIP}}$.
The local logits are computed as the cosine similarity, denoted $s(\cdot,\cdot)$, between the $v^{\texttt{FG}}$ token and the learnable classifier embeddings $W_{\text{LLM}}^*$, formalized as expressed in \cref{eq:local_logits}.
We then compute global logits using $v^{\texttt{CLS}}$, which captures the holistic image representation. Similarly to the local logits, in \cref{eq:global_logits} the global logits are obtained by projecting $v^{\texttt{CLS}}$ via $P_{\text{CLIP}}$ and compared against the same classifier embeddings to ensure semantic consistency. As our goal is to complement the global priors in the \texttt{[CLS]} token with fine-grained cues in \texttt{[FG]}, to produce the final logits, we fuse the local and global logits by computing their average, ensuring a symmetric representation. Finally, the symmetrically fused logits from the TokenFusion module are defined as given in \cref{eq:token_fusion}.
\begin{align}
\label{eq:local_logits}
\text{Logits}_{\text{local}} &= s(P_{\text{CLIP}}(v^{\texttt{FG}}) , W_{\text{LLM}}^*)\\
\label{eq:global_logits}
\text{Logits}_{\text{global}} &= s(P_{\text{CLIP}}(v^{\texttt{CLS}}), W_{\text{LLM}}^*)\\
\label{eq:token_fusion}
\text{TokenFusion}(x, W_{\text{LLM}}^*) &= \frac{\text{Logits}_{\text{local}} + \text{Logits}_{\text{global}}}{2}
\end{align}
We employ the same symmetric fusion framework during both training and inference, ensuring that global and fine-grained features receive equal supervision throughout self-training. This consistency encourages the model to learn complementary representations, leading to final predictions that reflect agreement between the \texttt{[FG]} and \texttt{[CLS]} tokens.

\label{sec:pseudo-labeling}
\subsection{Iteratively Improving Pseudo-Labels with Dynamic Knowledge Aggregation}

We build upon the core insight behind WCA~\cite{wca} but reconceptualize its components to enable a principled and efficient pseudo-labeling pipeline. Rather than treating a large number of localized crops as iterative ``visual prompts'' ($N\approx 60$), we model the multi-crop strategy as a weak augmentation $\alpha(x)$ and use a compact set of views to form a stable multi-view representation aligned with LLM-derived classifiers. This reframing drastically reduces computation, and via our Dynamic Knowledge Aggregation, provides a principled way to fuse static CLIP priors from multi-view alignment with the dynamically learned coarse- and fine-grained features in TokenFusion. Formally, for an unlabeled image $x \in \mathbb{R}^{H \times W \times 3}$, we generate $N$ random image crops:
\begin{equation}
\label{eq:multi_crop}
\alpha(x) = \left\{x_i | x_i = \phi(x, \lambda_i \min(H, W)) \mid i = 1\dots N \right\}
\end{equation}
where $\phi$ extracts a random crop of scale $\lambda_i \sim \mathcal{U}(a, b)$, and $\mathcal{U}(a, b)$ denotes the continuous uniform distribution over the interval $[a, b]$. We treat each crop $x_i$ as a weakly augmented view of the input image $x$, and extract its features using the CLIP vision encoder. 
To assess the relevance of each crop, a weight $w_i$ is computed by comparing its embedding with the global image embedding $v^{\texttt{CLS}}$~\cite{wca}: 
\begin{equation}
\label{eq:crop_weight}
w_i = \frac{\exp\left(s(f(x), f(x_i))\right)}{\sum_{l=1}^{N} \exp\left(s(f(x), f(x_l))\right)}
\end{equation}
In \cref{eq:crop_weight}, $f(x_i)$ denotes the CLIP embedding of the $i$-th crop, and $f(x) = P_{\text{CLIP}}(v^{\texttt{CLS}})$ is the global image representation.
We then aggregate the weighted crop embeddings to obtain a single representation, as expressed in \cref{eq:agg_img}. We interpret $f^{\text{agg}}(x)$ as an augmented visual representation that better aligns with a text-based fine-grained classifier since it emphasizes semantically rich local regions in the image. We align this aggregated representation with the fixed fine-grained textual classifier $W_{\text{LLM}}$, as given in \cref{eq:CLIP_logits}, to overcome the coarse-grained limitations of the \texttt{[CLS]} token and generate consistent pseudo-predictions that serve as a foundation for the next stage.
\begin{align}
\label{eq:agg_img}
f^{\text{agg}}(x) &= \sum_{i=1}^N w_i \cdot f(x_i | \alpha)\\
\label{eq:CLIP_logits}
\text{Pseudo-logits}_{\text{CLIP}} &= s(f^{\text{agg}}(x), W_{\text{LLM}} | \alpha)
\end{align}
Finally, to progressively refine pseudo-labels, we introduce \textit{Dynamic Knowledge Aggregation}, a mechanism that fuses pretrained knowledge from CLIP (via multi-crop alignment, \cref{eq:multi_crop}) with the dynamically evolving coarse- and fine-grained features learned by the TokenFusion module (\cref{eq:token_fusion}). We use learnable LLM-derived classifier embeddings, $W_{\text{LLM}}^*$, in the TokenFusion module to promote better alignment of fine-grained representations across both visual and textual modalities as given in \cref{eq:dyn_label}.
\begin{align}
\label{eq:dyn_label}
\hat{y} = \arg\max_{y \in \mathcal{Y}} \big\{
  \gamma \cdot \text{Pseudo-logits}_{\text{CLIP}} \nonumber \\
  +\; (1{-}\gamma) \cdot \text{TokenFusion}(x, W_{\text{LLM}}^*)
\big\}
\end{align} 
This aggregation enables the model to refine its predictions iteratively, enhancing label quality. We realize this aggregation as a convex combination, where the relative contribution of static and dynamic knowledge sources is modulated by a weighting coefficient $\gamma$.
During training, a strongly augmented version of the target image, denoted $\mathcal{A}(x)$, is used and supervised by $\hat{y}$. The corresponding loss is a cross-entropy objective as expressed in \cref{eq:self-training}.  
\begin{align}
    \label{eq:self-training}
    \mathcal{L}_{\text{st}} &= - \; \mathbb{E}_{x \in \mathcal{X}_t} \; \sum_{j=1}^C  \!\mathbb{I}\{\hat{y}=j \!\}\! \; log \left(\text{TF}(\mathcal{A}(x),W_{\text{LLM}}^*)\right)
\end{align}
Here, $\text{TF}(\cdot,\cdot)$ represents the TokenFusion module.  
To mitigate confirmation bias and class imbalance issues commonly encountered in CLIP adaptation~\cite{li2022masked, Wang2022DebiasedLF}, we further incorporate a fairness regularization loss inspired by~\cite{li2022masked}:$\mathcal{L}_{\text{reg}} = -\frac{1}{C} \sum_{k=1}^C \log \bar{p}_{\mathcal{A}(x^k)}$, where $\bar{p}_{\mathcal{A}(x^k)}$ denotes the average predicted probability over a mini-batch for class $k$. This regularization promotes a uniform prediction distribution across classes, thereby reducing overfitting to noisy pseudo-labels and encouraging balanced adaptation. The overall loss function used for training is: $\mathcal{L} = \mathcal{L}_{\text{st}} + \mathcal{L}_{\text{reg}}$. 

\section{Experiments and Analyses}

\begin{table*}[t]
    \centering
    \scriptsize
    \setlength{\tabcolsep}{3pt} 
    \renewcommand{\arraystretch}{1.1} 
    \begin{tabularx}{\textwidth}{l >{\centering\arraybackslash}p{1cm} *{14}{>{\centering\arraybackslash}X}}
      \toprule
      \textbf{Method} & \textbf{Venue} & 
      \rotatebox{90}{\textbf{Birdsnap}} & 
      \rotatebox{90}{\textbf{Caltech}} & 
      \rotatebox{90}{\textbf{Cars}} & 
      \rotatebox{90}{\textbf{CIFAR100}} & 
      \rotatebox{90}{\textbf{DTD}} & 
      \rotatebox{90}{\textbf{FGVC}} & 
      \rotatebox{90}{\textbf{Flowers}} & 
      \rotatebox{90}{\textbf{Food101}} & 
      \rotatebox{90}{\textbf{Imagenet}} & 
      \rotatebox{90}{\textbf{Pets}} & 
      \rotatebox{90}{\textbf{RESISC}} & 
      \rotatebox{90}{\textbf{SUN397}} & 
      \rotatebox{90}{\textbf{UCF101}} & 
      \rotatebox{90}{\textbf{Avg}} \\
      \midrule
      
      \rowcolor[gray]{0.9} \multicolumn{16}{c}{\textbf{Zero-shot / Training-free Methods}} \\
      \midrule
      CLIP~\cite{clip} & ICML'21 & 37.45 & 90.69 & 58.70 & 64.47 & 44.63 
      & 19.50 & 66.42 & 83.95 & 63.30 & 87.50 & 57.59 & 61.32 & 61.86 & 61.34 \\
      CuPL~\cite{CuPL} & ICCV'23 & 37.02 & {94.62} & 60.79 & 65.22 & 50.11 
      & 20.94 & 69.51 & 84.05 & 64.26 & 87.16 & 61.14 & 65.57 & 66.90 & 63.64 \\ 
    WCA$^{*}$~\cite{wca} & ICML'24 & 37.63 & 94.02 & \underline{61.95} & 51.78 & 51.60 & \underline{21.15} & 68.70 & 83.97 & \textbf{65.01} & 86.32 & 62.56 & 64.93 & 65.82 & 62.73\\

      \midrule
      \rowcolor[gray]{0.9} \multicolumn{16}{c}{\textbf{UA Methods}} \\
      \midrule
      
    UPL~\cite{upl} & - & 32.80 & 92.36 & 49.41 & 67.41 & 45.37 
    & 17.07 & 67.40 & 84.25 & 58.22 & 83.84 & 57.63 & 62.12 & 62.04 & 59.99 \\
    POUF~\cite{pouf} & ICML'23 & \underline{38.40} & 94.10 & 57.70 & 62.00 & 46.10 
    & 18.20 & 67.80 & 82.10 & 52.20 & 87.80 & 66.40 & 60.00 & 61.20 & 61.08 \\
    LaFTer~\cite{lafter} & NeurIPS'23 & 21.14 & 94.39 & 57.44 & 69.79 & 50.32 
    & 19.86 & 72.43 & 82.45 & 61.63 & 84.93 & 61.60 & 65.87 & 65.08 & 62.07 \\
    ReCLIP$^{\dagger}$~\cite{reclip} & WACV'24 & 37.38 & 93.84 & 58.84 & 71.43 & 53.88 
    & 18.87 & 72.63 & 84.22 & 63.95 & 85.27 & \underline{73.05} & 65.23 & \underline{67.06} & 64.69 \\
    DPA$^{\ddagger}$~\cite{DPA} & WACV'25 & 31.54 & \textbf{95.54} & 56.83 & \underline{74.22} & \underline{55.96} 
    & 20.10 & \underline{75.48} & \underline{84.76} & \underline{64.64} & \underline{90.11} & 71.11 & \underline{68.13} & 66.69 & \underline{65.78} \\
    \midrule
      
    \rowcolor[HTML]{FFD6E7} \textbf{\ours} (Ours) & - & \textbf{38.59} & \underline{94.93} &	\textbf{65.81} & \textbf{77.41} & \textbf{60.00} 
    &	\textbf{22.74} &	\textbf{75.84} &	\textbf{85.58} &	64.45 & \textbf{90.24} &	\textbf{77.25} & \textbf{68.98} &	\textbf{70.98} &	\textbf{68.68} \\ 
      
      \bottomrule
    \end{tabularx}
    
  \caption{Top-1 accuracy (\%) comparison for 13 datasets of state-of-the-art methods using the ViT-B/32 backbone. $*$ represents the reproduced results using the same number of crops as \ours. $^{\dagger}$ We get the results by training ReCLIP~\cite{reclip} under inductive settings. 
  $^{\ddagger}$ For fair comparison, we reproduce DPA using the same fixed learning rate as \ours.
  }
  \label{table:sota_vit_b_32}
\end{table*}

\textbf{Datasets and Training Setup:} We evaluate \ours on 13 varied datasets: Birdsnap~\cite{birdsnap}, Caltech~\cite{fei2004learning}, Cars~\cite{krause20133d}, CIFAR100~\cite{krizhevsky2009learning}, DTD~\cite{cimpoi2014describing}, FGVC~\cite{maji2013fine}, Flowers~\cite{nilsback2008automated}, Food101~\cite{bossard2014food}, ImageNet~\cite{deng2009imagenet}, Pets~\cite{parkhi2012cats}, RESISC~\cite{Cheng2017RemoteSI}, SUN397~\cite{xiao2010sun}, and UCF101~\cite{soomro2012ucf101}. These datasets span diverse domains, supporting a thorough evaluation of generalization.
We benchmark our method against eight state-of-the-art approaches, including zero-shot methods such as CLIP~\cite{clip}, CuPL~\cite{CuPL}, and WCA~\cite{wca}, as well as unsupervised adaptation techniques for CLIP: UPL~\cite{upl}, POUF~\cite{pouf}, LaFTer~\cite{lafter}, ReCLIP~\cite{reclip}, and DPA~\cite{DPA}. All experiments utilize a ViT/B-32 CLIP model pretrained by OpenAI~\cite{clip}. During fine-tuning, we adopt the approach from~\cite{reclip, DPA} to adjust only the layer normalization weights of the image encoder~\cite{ba2016layer}, improving stability under noisy supervision~\cite{wang2020tent}, while also fine-tuning the text-based classifier embeddings $W_{\text{LLM}}^*$. For details on the construction of $W_{\mathrm{LLM}}$ and $W_{\mathrm{LLM}}^*$, please refer to the Supp. A.1.
Based on ablation experiments reported in the results section, we set $\gamma = 0.5$ and use $N = 8$ for multi-view alignment in the pseudo-labeler across all datasets. Our learning rate policy and its sensitivity analysis appear in Supp. B.1 (Fig. 2).

\subsection{Main Results}

We report overall accuracy across 13 fine-grained datasets in \cref{table:sota_vit_b_32}. \ours consistently outperforms both zero-shot and UA baselines that rely on CLIP’s coarse-grained representations, using the ViT-B/32 backbone. Compared to the strongest prior UA method, DPA, \ours achieves an overall accuracy of $68.68\%$, setting a new state-of-the-art with a $2.90\%$ gain. Notably, our method yields substantial improvements on FGVC ($+2.64\%$), a dataset that is particularly challenging in unsupervised settings. It also demonstrates strong gains on several benchmarks, including Cars ($+8.98\%$), RESISC ($+6.14\%$), UCF101 ($+4.29\%$), CIFAR100 ($+3.19\%$), and DTD ($+4.04\%$). It is worth highlighting that UA methods have historically struggled with Cars due to their fine-grained intra-class variations and high inter-class similarity; yet \ours surpasses the best-performing UA method on Cars (ReCLIP) by $+6.97\%$. See Supp. B.2 (Tab. 4) for 1–2-shot comparisons and Supp. B.3 (Tab. 5) for comparisons on additional VLMs. Limitations of our method are discussed in the Supp. E.

\subsection{Ablation Studies}
\textbf{Naive Coarse-feature Fine-tuning Baselines:}
\label{Naive_Global_Fine-tuning_Baselines}
\Cref{tab:global_finetuning} highlights the critical importance of incorporating fine-grained cues for fine-tuning. Compared to two baselines differing in the visual representation (single-view vs. multi-view) used to generate pseudo-labels (PL), where only the \texttt{[CLS]} token is aligned with the learnable classifier ($W_{\text{LLM}}^*$) during training, our approach achieves a notable improvement of $2.40\%$ over the best-performing baseline. We validate this through two PL setups for fairness: (1) using fixed classifier embeddings ($W_{\text{LLM}}$) for PL, and (2) a shared classifier setting where $W_{\text{LLM}} = W_{\text{LLM}}^*$. In both cases, results consistently show that relying solely on the \texttt{[CLS]} token leads to suboptimal performance, underscoring the necessity of the proposed \texttt{[FG]} token. 

\begin{table}[!ht]
  \centering
  \scriptsize
  \setlength{\tabcolsep}{3pt}
  \renewcommand{\arraystretch}{1.15}
  \begin{tabularx}{\linewidth}{>{\raggedright\arraybackslash}p{3cm} *{7}{>{\centering\arraybackslash}X}}
    \toprule
    \textbf{Component} & 
    \rotatebox{90}{\textbf{Cars}} & 
    \rotatebox{90}{\textbf{DTD}} & 
    \rotatebox{90}{\textbf{FGVC}} & 
    \rotatebox{90}{\textbf{Flowers}} &
    \rotatebox{90}{\textbf{Pets}} & 
    \rotatebox{90}{\textbf{UCF101}} & 
    \rotatebox{90}{\textbf{Avg}} \\
     \midrule
    \rowcolor[HTML]{E6E6E6} \multicolumn{8}{c}{\textbf{Fixed Classifier Embeddings for PL}} \\
    \midrule
    Single-view Alignment PL & 61.95 & 53.72 & \underline{21.96} & 72.51 & 89.18 & 68.86 & 61.36 \\
    Multi-view Alignment PL & \underline{63.28}	& 55.96	& 21.72	& 72.35	& 88.69	& 69.23	& \underline{61.87} \\
    \midrule
    \rowcolor[HTML]{E6E6E6} \multicolumn{8}{c}{\textbf{Shared Learnable Classifier Embeddings for PL}} \\
    \midrule
    Single-view Alignment PL & 56.81 & 59.10 & 16.26 & \underline{72.67} & \underline{89.78} & 70.16 & 60.80 \\
    Multi-view Alignment PL & 56.01 & \textbf{61.76} & 11.31 & 72.31 & 90.24 & \textbf{71.95} & 60.60 \\
    \midrule
    \rowcolor[HTML]{FFD6E7} \textbf{\ours} (Ours) & \textbf{65.81} & \underline{60.00} & \textbf{22.74} & \textbf{75.84} & \textbf{90.24} & \underline{70.98} & \textbf{64.27} \\
    \bottomrule
  \end{tabularx}
  \caption{\textnormal{Ablation on coarse-feature fine-tuning baselines.}}
    \label{tab:global_finetuning}
\end{table}

\noindent\textbf{Saliency-Oriented Attention Pooling:} We assess SOAP’s impact in~\cref{tab:attention_pooling}. Replacing it with naive token averaging for \texttt{[FG]} leads to a $1.97\%$ drop in average accuracy. Using the average of NCut selection only results in $60.71\%$, likely because averaging disregards the relative importance and saliency of the selected tokens, thereby diluting the focus on discriminative features. Since SOAP relies on a saliency-aware query, we test two weaker alternatives: (i) naive token averaging, and (ii) random token selection, resulting in $1.71\%$ and $1.39\%$ drops, respectively. These queries fail to emphasize semantically relevant regions, unlike NCut, which selects the most coherent and salient tokens for more discriminative attention. \Cref{fig:teaser_motivation} (bottom row) and Fig. 3 (supplementary) further provide qualitative evidence supporting SOAP’s effectiveness.

\begin{table}[!ht]
  \centering
  \scriptsize
  \setlength{\tabcolsep}{3pt}
  \renewcommand{\arraystretch}{1.15}
  \begin{tabularx}{\linewidth}{>{\raggedright\arraybackslash}p{3.5cm} *{7}{>{\centering\arraybackslash}X}}
    \toprule
    \textbf{Component} & 
    \rotatebox{90}{\textbf{Cars}} & 
    \rotatebox{90}{\textbf{DTD}} & 
    \rotatebox{90}{\textbf{FGVC}} & 
    \rotatebox{90}{\textbf{Flowers}} &
    \rotatebox{90}{\textbf{Pets}} & 
    \rotatebox{90}{\textbf{UCF101}} & 
    \rotatebox{90}{\textbf{Avg}} \\
     \midrule
    \rowcolor[HTML]{E6E6E6} \multicolumn{8}{c}{{\textbf{No Attention Pooling}}} \\
    \midrule
   Naive Token Average as \texttt{[FG]} & 63.23 & 57.61 & 18.72 & 74.30 & 89.13 & \underline{70.82} & 62.30 \\
   NCut Token Average as \texttt{[FG]} & 59.21 & 56.54 & 17.49 & 73.24 & 88.36 & 69.42 & 60.71 \\
      \midrule
  \rowcolor[gray]{0.9} \multicolumn{8}{c}{\textbf{Attention Pooling Query}} \\
      \midrule
    Naive Token Average & 62.83 & \underline{58.56} & \underline{21.15} & 73.45 & 89.23 & 70.13 & 62.56 \\
    Random Token Selection & \underline{63.89} & 58.03 & 19.86 & \textbf{76.17} & \underline{89.83} & 69.52 & \underline{62.88} \\
      \midrule
    \rowcolor[HTML]{FFD6E7} \textbf{SOAP} (Ours) & \textbf{65.81} & \textbf{60.00} 
    & \textbf{22.74} & \underline{75.84} & \textbf{90.24} & \textbf{70.98} & \textbf{64.27} \\
      \bottomrule
  \end{tabularx}
  \caption{\textnormal{Ablation on Attention Pooling.}}
    \label{tab:attention_pooling}
\end{table}

\begin{figure}[t]
    \centering
    \includegraphics[width=0.9\columnwidth]{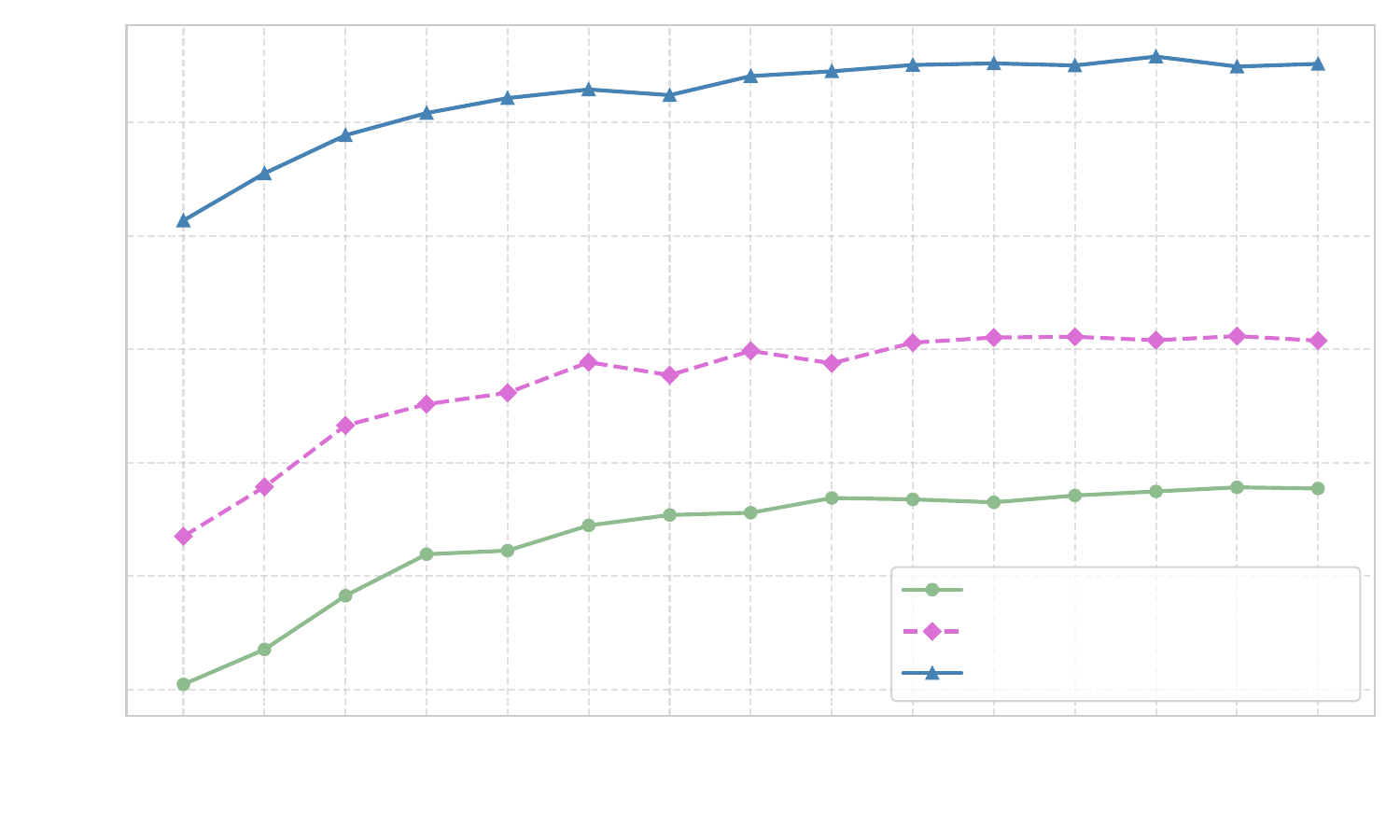}
    \caption{Pseudo-Labeling Accuracy variation of each component and Dynamic Knowledge Aggregation over time on the Stanford Cars train split.}
    \label{fig:pl_accuracy}
\end{figure}

\noindent\textbf{Pseudo Labeler:} 
In \cref{tab:pseudo_labeling}, we validate the effectiveness of our \textit{Dynamic Knowledge Aggregation} strategy through an ablation study on pseudo-labeling (PL) classifier configurations. Under the previously mentioned single-view and multi-view alignment setups, performance drops by $3.61\%$ and $3.11\%$ when the fine-tuned classifier shares parameters with the pseudo-labeler. In contrast, using only fixed classifiers results in relatively smaller drops of $1.57\%$ and $1.48\%$. Moreover, removing pretrained knowledge from PL generation ($\gamma = 0$) leads to a sharp decline in performance to $58.03\%$. These results demonstrate that both pretrained and newly learned knowledge are essential; neither alone is sufficient. \Cref{fig:pl_accuracy} shows the progression of PL accuracy over training epochs on the Cars dataset. While `Multi-view Alignment Only ($\gamma=0$)' PL (\cref{eq:CLIP_logits}) accuracy remains relatively stagnant, `\ours Only ($\gamma=1$)' PL accuracy steadily improves, reflecting the advantage of capturing fine-grained visual cues.  Notably, `Dynamic Aggregation ($\gamma=0.5$)' achieves higher pseudo-label accuracy when knowledge from both sources is aggregated during training, underscoring the benefit of fusing static and dynamic supervision.

\begin{table}[!ht]
  \centering
  \scriptsize
  \setlength{\tabcolsep}{3pt}
  \renewcommand{\arraystretch}{1.15}
  \begin{tabularx}{\linewidth}{>{\raggedright\arraybackslash}p{3.5cm} *{7}{>{\centering\arraybackslash}X}}
    \toprule
    \textbf{Component} & 
    \rotatebox{90}{\textbf{Cars}} & 
    \rotatebox{90}{\textbf{DTD}} & 
    \rotatebox{90}{\textbf{FGVC}} & 
    \rotatebox{90}{\textbf{Flowers}} &
    \rotatebox{90}{\textbf{Pets}} & 
    \rotatebox{90}{\textbf{UCF101}} & 
    \rotatebox{90}{\textbf{Avg}} \\
     \midrule
    \rowcolor[gray]{0.9} \multicolumn{8}{c}{\textbf{Fixed Classifier Embeddings for PL}} \\
      \midrule
    Single-view Alignment PL & 64.81 & 55.80 
    & 22.47 & \underline{73.69} & 89.32 & \underline{70.08} & 62.69 \\
    Multi-view Alignment PL ($\gamma=1$) & \underline{65.10} & 56.76 & \textbf{23.01} & 73.57 & 88.55 & 69.76 & \underline{62.79} \\
    \midrule
    \rowcolor[gray]{0.9} \multicolumn{8}{c}{\textbf{Shared Learnable Classifier Embeddings for PL}} \\
    \midrule
    Single-view Alignment PL & 60.27 & 57.71 
    & 15.48 & 72.59 & 88.91 & 68.97 & 60.66 \\
    Multi-view Alignment PL & 59.91 & \underline{59.95} & 16.56 & 72.84 & 88.39 & 69.28 & 61.16 \\
    TokenFusion Logits Only ($\gamma = 0$) & 55.34 &	54.36 &	10.14 &	70.40 &	\underline{89.34} &	68.62 & 58.03 \\
    \midrule
    \rowcolor[HTML]{FFD6E7} \textbf{Dynamic Knowledge Aggregation} (Ours) & \textbf{65.81} & \textbf{60.00} 
    & \underline{22.74} & \textbf{75.84} & \textbf{90.24} & \textbf{70.98} & \textbf{64.27} \\
      \bottomrule
  \end{tabularx}
  \caption{Ablation on the pseudo-labeler.}
  \label{tab:pseudo_labeling}
\end{table}

\noindent\textbf{Two-headed Classifier:} To evaluate the impact of text prompt initialization for our two classifiers, $W_{\text{LLM}}$ and $W_{\text{LLM}}^{*}$, we conduct an ablation study on various strategies, as detailed in \cref{tab:classifier_ablation}. We employ the same prompt ensembling technique as CLIP~\cite{clip} for class-specific handcrafted prompts. Consistent with WCA's design choices~\cite{wca}, we exclude the ablation where handcrafted prompts are used for a fixed $W_{\text{LLM}}$. The results demonstrate that our method achieves superior performance across the ablation datasets, with overall accuracy gains of $2.03\%$ and $1.99\%$ compared to the two ablation settings.
\begin{table}
  \centering
    \scriptsize
    \setlength{\tabcolsep}{3pt} 
    \begin{tabularx}{\linewidth}{l >{\centering\arraybackslash} *{7}{>{\centering\arraybackslash}X}}
      \toprule
      \textbf{Component} & 
    \rotatebox{90}{\textbf{Cars}} & 
    \rotatebox{90}{\textbf{DTD}} & 
    \rotatebox{90}{\textbf{FGVC}} & 
    \rotatebox{90}{\textbf{Flowers}} &
    \rotatebox{90}{\textbf{Pets}} & 
    \rotatebox{90}{\textbf{UCF101}} & 
    \rotatebox{90}{\textbf{Avg}} \\
      \midrule
    Handcrafted prompts for $W_{\text{LLM}}^*$ & \underline{65.08} & \underline{58.98} 
    & \underline{19.95} & 69.50 & 89.97 & \underline{69.97} & 62.24 \\
    Handcrafted prompts for both & {64.32} & {57.07} 
    & 19.05 & \underline{74.26} & \underline{90.11} & {68.86} & \underline{62.28} \\
    \midrule
    \rowcolor[HTML]{FFD6E7} \textbf{LLM descriptions for both} (Ours) & \textbf{65.81} & \textbf{60.00} 
    & \textbf{22.74} & \textbf{75.84} & \textbf{90.24} & \textbf{70.98} & \textbf{64.27} \\
      \bottomrule
    \end{tabularx}
  \caption{Ablation on the Two-headed Classifier.}
  \label{tab:classifier_ablation}
\end{table}
\begin{table}
  \centering
    \scriptsize
    \setlength{\tabcolsep}{3pt} 
    \begin{tabularx}{\linewidth}{l >{\centering\arraybackslash} *{7}{>{\centering\arraybackslash}X}}
      \toprule
      \textbf{Component} & 
    \rotatebox{90}{\textbf{Cars}} & 
    \rotatebox{90}{\textbf{DTD}} & 
    \rotatebox{90}{\textbf{FGVC}} & 
    \rotatebox{90}{\textbf{Flowers}} &
    \rotatebox{90}{\textbf{Pets}} & 
    \rotatebox{90}{\textbf{UCF101}} & 
    \rotatebox{90}{\textbf{Avg}} \\
      \midrule
    \rowcolor[gray]{0.9} \multicolumn{8}{c}{\textbf{Zero-shot Methods}} \\
  \midrule
    CLIP~\cite{clip} & 64.70 & 44.70 
    & 23.97 & 70.89 & 89.00 & 69.10 & 60.39 \\
    CuPL~\cite{CuPL} & \underline{64.92} & 53.46 
    & \underline{27.72} & 73.37 & 90.71 & 69.42 & 63.27 \\
    \midrule
    \rowcolor[gray]{0.9} \multicolumn{8}{c}{\textbf{UA Methods}} \\
    \midrule
    UPL~\cite{upl} & 60.33 & 45.90 
    & 22.53 & 73.93 & 87.98 & 67.43 & 59.68 \\
    POUF~\cite{pouf} & 63.50 & 48.60 
    & 24.40 & 72.10 & 91.80 & 71.50 & 61.98 \\
    LaFTer~\cite{lafter} & 64.72 & \underline{54.79} 
    & 22.38 & 75.15 & 85.28 & 67.20 & 61.59 \\
    DPA~\cite{DPA} & 63.97 & 50.32 
    & 20.10 & \underline{78.64} & \underline{93.35} & \underline{74.44} & \underline{63.47} \\
    \midrule
    \rowcolor[HTML]{FFD6E7} \textbf{\ours} (Ours) & \textbf{72.50} & \textbf{60.74} 
    & \textbf{31.29} & \textbf{79.86} & \textbf{93.43} & \textbf{75.18} & \textbf{68.83} \\
      \bottomrule
    \end{tabularx}
  \caption{Top-1 accuracy (\%) comparison using the ViT-B/16 backbone.}
  \label{tab:vit_b16}
\end{table}

\noindent \textbf{Ablation on Token Fusion:}
We conduct an ablation by removing the fusion in~\cref{eq:token_fusion}, using only one of the two components. \ours normally averages the global \texttt{[CLS]} token logits and local patch token logits to balance coarse and fine-grained cues. We test two variants: (i) global-only and (ii) local-only. As shown in~\cref{tab:local_global}, the global-only model performs poorly (17.26\%), while the local-only variant does better (57.84\%), highlighting the importance of fine-grained features. Still, both fall short of our full method, confirming that combining global and local cues is crucial for robust pseudo-labeling.

\begin{table}[!ht]
  \centering
  \scriptsize
  \setlength{\tabcolsep}{3pt}
  \renewcommand{\arraystretch}{1.15}
  \begin{tabularx}{\linewidth}{>{\raggedright\arraybackslash}p{3cm} *{7}{>{\centering\arraybackslash}X}}
    \toprule
    \textbf{Component} & 
    \rotatebox{90}{\textbf{Cars}} & 
    \rotatebox{90}{\textbf{DTD}} & 
    \rotatebox{90}{\textbf{FGVC}} & 
    \rotatebox{90}{\textbf{Flowers}} &
    \rotatebox{90}{\textbf{Pets}} & 
    \rotatebox{90}{\textbf{UCF101}} & 
    \rotatebox{90}{\textbf{Avg}} \\
     \midrule
    \rowcolor[gray]{0.9} \multicolumn{8}{c}{\textbf{Fixed Classifier Embeddings for PL}} \\
      \midrule
    Global logits only & 5.71&	34.41	&2.19&	24.20	&30.93&	6.13	&17.26 \\
    Local logits only & \underline{60.05}	& \underline{52.29}	& \underline{21.72}	&\underline{60.63}	&\underline{86.35}&	\underline{65.98}	&\underline{57.84} \\
    \midrule
    \rowcolor[HTML]{FFD6E7} \textbf{Symmetric Fusion} (Ours) & \textbf{65.81} & \textbf{60.00} & \textbf{22.74} & \textbf{75.84} & \textbf{90.24} & \textbf{70.98} & \textbf{64.27} \\
      \bottomrule
  \end{tabularx}
  \caption{Ablation on TokenFusion Symmetry.}
  \label{tab:local_global}
\end{table}

\noindent\textbf{ViT-B/16 Backbone:} Using ViT-B/16 as the CLIP backbone, our method outperforms prior approaches (\cref{tab:vit_b16}), with a $5.36\%$ gain over DPA. This substantial gain is attributed to the smaller patch size of ViT-B/16, which yields richer fine-grained patch tokens for our SOAP.

\begin{figure}[t]
    \centering
    \includegraphics[width=\columnwidth]{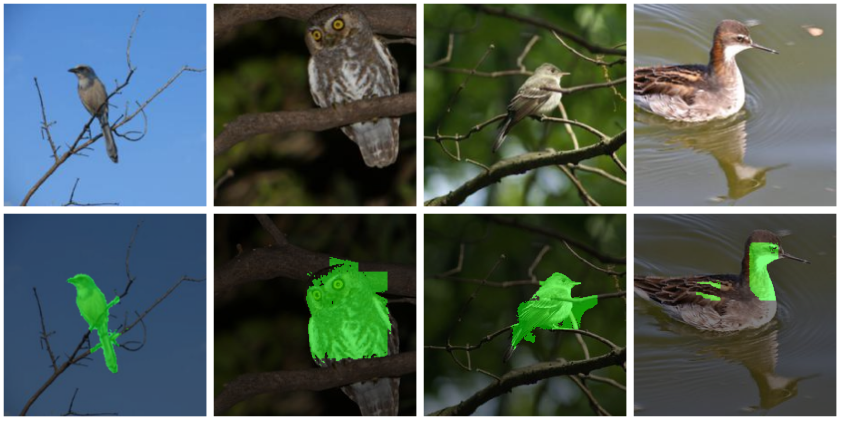}
    \caption{NCut-based saliency masks on bird images from Birdsnap~\cite{birdsnap}. Top: input images; bottom: salient regions after CRF refinement.}
    \label{fig:ncut_visualization}
\end{figure}

\noindent \textbf{Saliency-based Region Extraction with NCut:}
We visualize the bipartition mask produced by our NCut-based saliency mechanism in~\cref{fig:ncut_visualization}. For visualization, we upsample and interpolate the NCut output and apply a Conditional Random Field (CRF) following~\cite{TokenCut}. The NCut of patch tokens consistently highlights object-centric regions across diverse bird images. 
This saliency awareness plays a key role in our SOAP query by enabling \emph{locally prompted} attention pooling.

\begin{figure}[!ht]
    \centering
    \includegraphics[width=0.8\columnwidth]{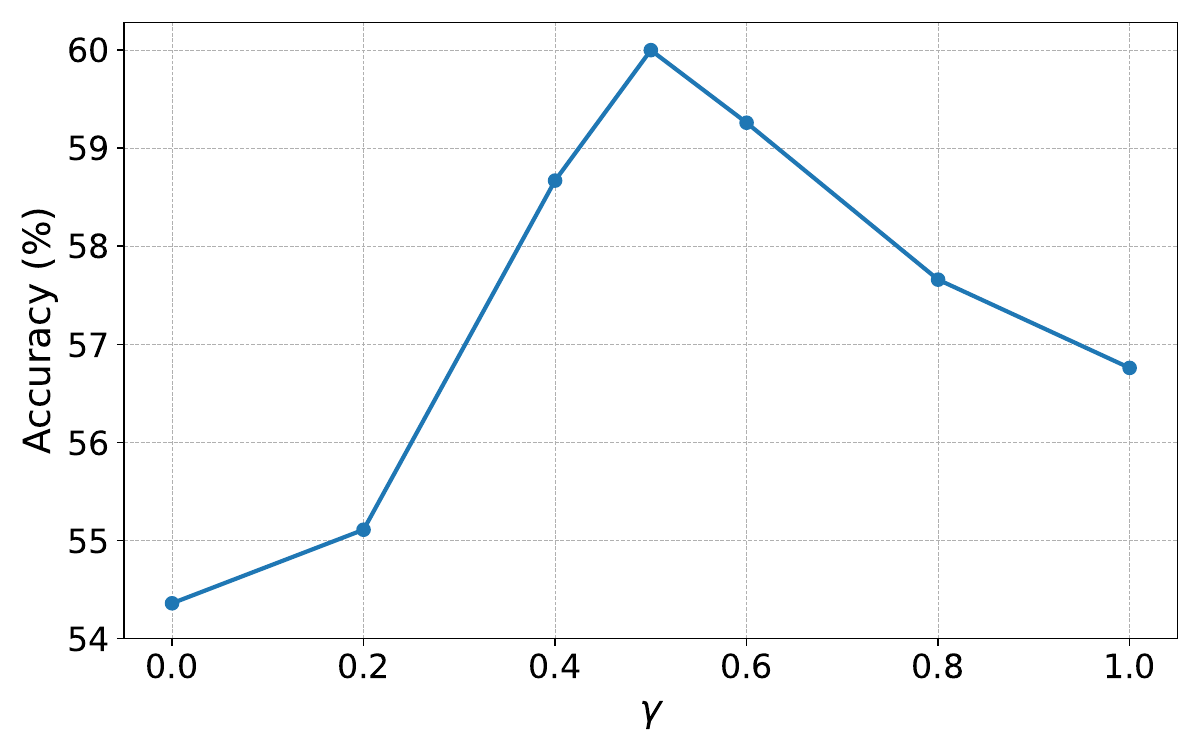}
    \caption{$\gamma$ sensitivity analysis on the DTD dataset.}
    \label{fig:gamma}
\end{figure}

\noindent \textbf{Sensitivity to $\gamma$:} 
We ablate the knowledge weighting coefficient $\gamma$ of Dynamic Knowledge Aggregation on DTD (\cref{fig:gamma}). Accuracy peaks at $60.00\%$ when $\gamma=0.5$, suggesting moderate values balance performance, while large ones cause instability or over-regularization.

\begin{figure}[!ht]
    \centering
    \begin{subfigure}[t]{0.47\linewidth}
        \centering
        \includegraphics[width=\linewidth]{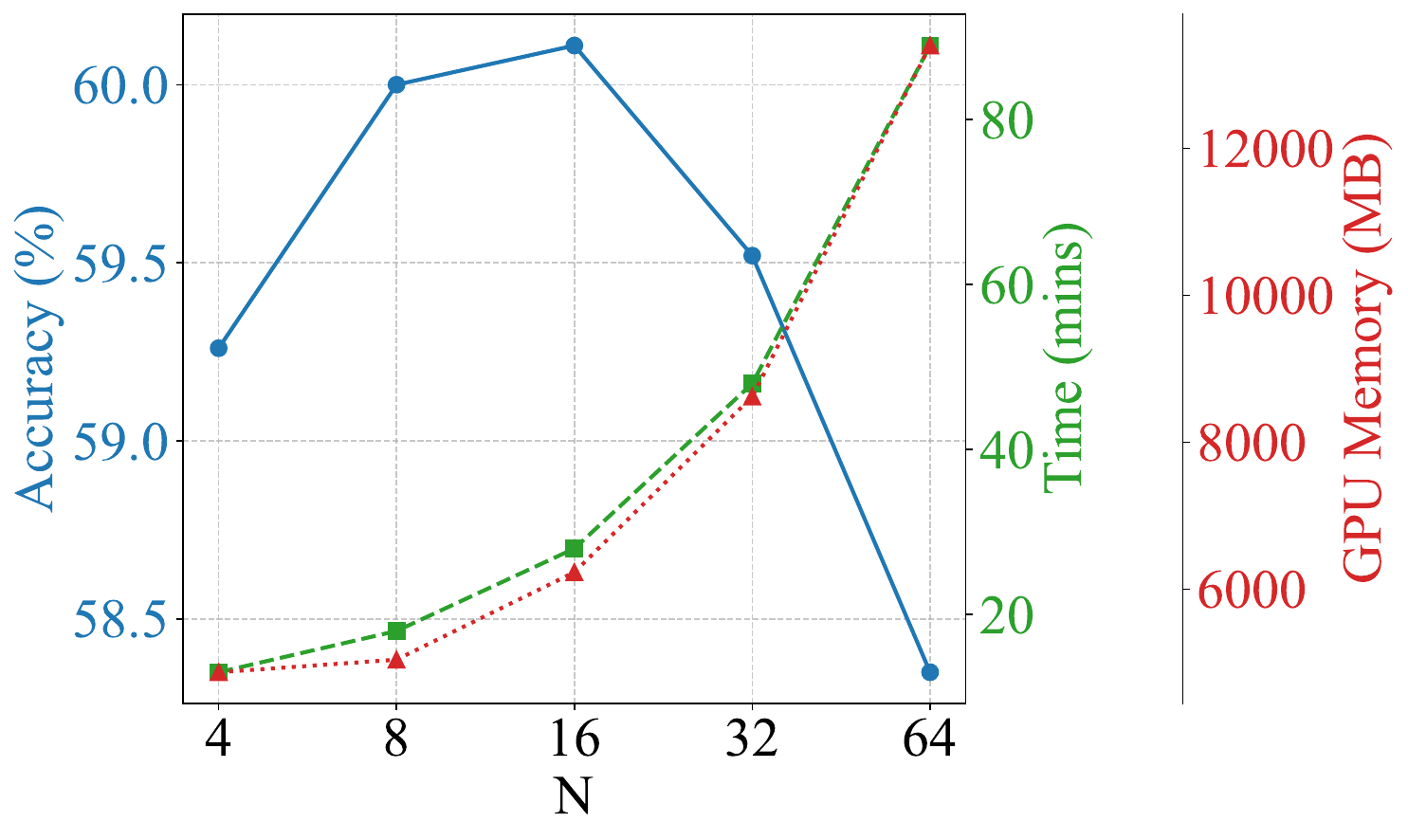}
        \caption{DTD dataset.}
        \label{fig:crops_dtd}
    \end{subfigure}
    \hfill
    \begin{subfigure}[t]{0.47\linewidth}
        \centering
        \includegraphics[width=\linewidth]{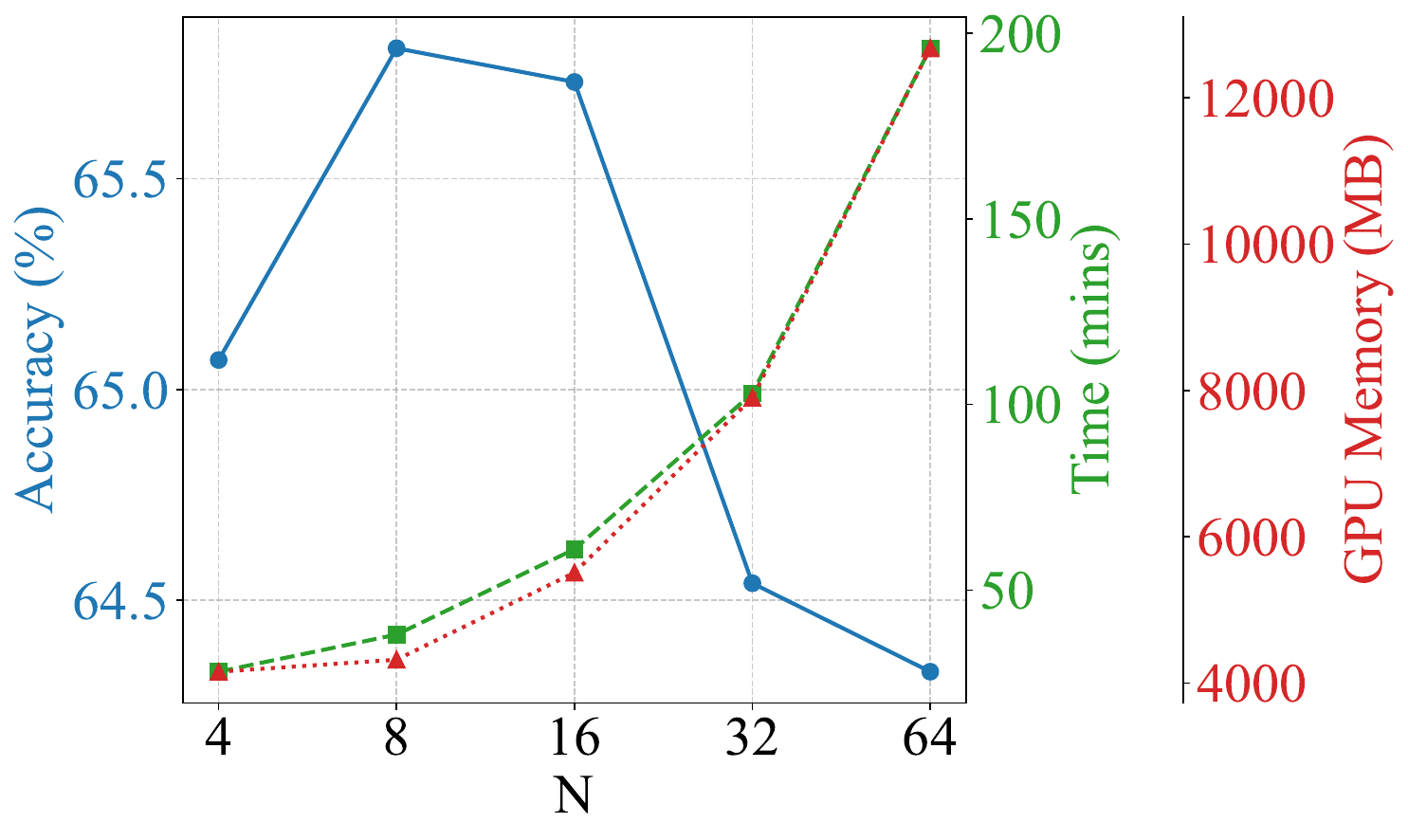}
        \caption{Cars dataset.}
        \label{fig:crops_cars}
    \end{subfigure}
    \caption{Analysis of accuracy, training time, and GPU memory usage across varying sampled crop sizes ($N$).}
    \label{fig:crops}
\end{figure}

\noindent \textbf{Numbers of Crops}:
We evaluate the impact of the number of image crops ($N$) on \ours performance using the DTD dataset, as shown in~\cref{fig:crops_dtd}. As $N$ increases, training time and GPU usage rise significantly. Accuracy peaks at $60.00\%$ with 8 crops and $60.11\%$ with 16 crops. Due to the marginal improvement, we select 8 crops to strike a balance between accuracy and resource efficiency. Further increasing $N$ leads to declining accuracy, while training time and GPU usage continue to rise.
To validate this, we conduct a similar analysis on the Cars dataset, as shown in~\cref{fig:crops_cars}. Accuracy peaks at $65.81\%$ with 8 crops, while training time and GPU usage increase with higher $N$. Thus, $N=8$ consistently provides the optimal balance between accuracy and computational efficiency.

\section{Conclusion}
\label{Conclusion}
We show that CLIP’s pretrained global \texttt{[CLS]} representation, while stable, can be insufficient for fine-grained unsupervised adaptation because many subtle distinctions rely on localized cues. To address this, we propose \ours, which augments CLIP with a fine-grained \texttt{[FG]} token obtained via Saliency-Oriented Attention Pooling (SOAP) inside a lightweight TokenFusion module, and aligns coarse and fine signals with a two-headed LLM-derived classifier: a frozen prototype \(W_{LLM}\) as a stable prior and an adaptive prototype \(W_{LLM}^*\) fine-tuned with TokenFusion. Rather than discarding the \texttt{[CLS]} token, we use it together with \(W_{LLM}\) to form stable pseudo-labels via multi-view alignment and refine them with Dynamic Knowledge Aggregation, convexly blending static priors and evolving TokenFusion logits. Empirically, \ours uncovers CLIP’s latent fine-grained cues and raises average accuracy on 13 fine-grained benchmarks from \(61.34\%\) to \(68.68\%\), establishing a new state-of-the-art. Overall, \ours is an effective and lightweight strategy for unsupervised fine-grained adaptation of CLIP.


\section*{Supplementary Material for \includegraphics[height=1em]{images/introduction/microscope_1f52c.png} microCLIP}

\appendix

We organize the supplementary materials into six appendices. 
In section~\cref{sec:additional_impl}, we detail additional implementation and technical aspects omitted from the main paper. In~\cref{sec:addition_exp}, we present additional experiments that further validate our results and provide insights into the motivations behind our design choices. \Cref{sec:qualitative_analysis} offers a visual analysis of our method by comparing global and local attention patterns from the \texttt{[CLS]} and \texttt{[FG]} tokens with the distinguishing features in the corresponding images, further demonstrating the effectiveness of our approach in capturing critical local semantics. In \cref{sec:ncut}, we provide the derivation of the Normalized Cut algorithm used to construct the saliency-oriented query for attention pooling. 
\cref{sec:limitations} discusses the limitations of our work and outlines potential future directions. Lastly, \cref{sec:pseudo_code} provides a summary of the symbols and notations, along with a pseudocode representation of our method.

\section{Additional Implementation and Technical Details}
\label{sec:additional_impl}

\subsection{Two-headed LLM-derived Classifier}

\begin{figure}[!ht]
    \centering
    \includegraphics[width=\linewidth]{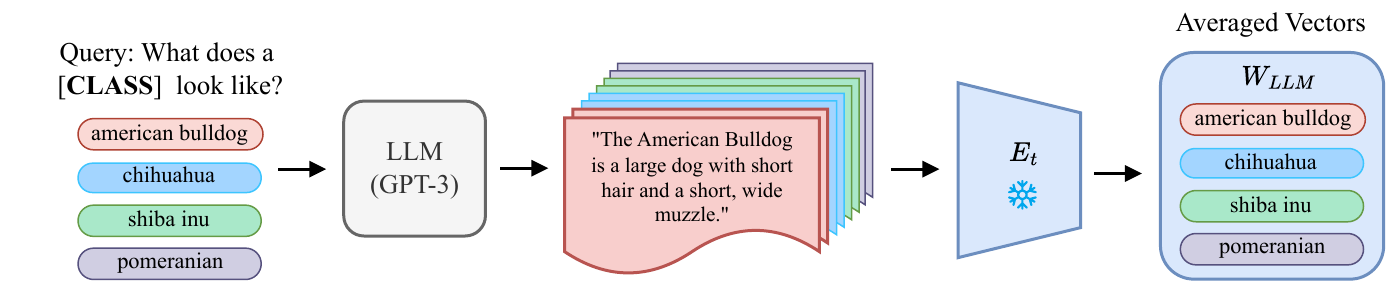}
    \caption{Initialization of the LLM-derived Classifiers.}
    \label{fig:llm_descriptions}
\end{figure}

For all experiments in the main paper that utilize LLM-derived classifiers, the descriptions used to construct $W_{\text{LLM}}$ and $W_{\text{LLM}}^{*}$ are sourced from CuPL~\cite{CuPL}. CuPL generates class-specific descriptions using two configurations, base and full, by carefully prompting a large language model (LLM). In the base configuration, three general handcrafted templates are used, such as \textcolor{darkbrown}{“Describe what a/the \{CLASS\} looks like.”}. In contrast, the full configuration employs dataset-specific prompts tailored to each dataset. As reported in CuPL, the full setting produces higher-quality descriptions that lead to better zero-shot performance due to the use of more context-aware prompting. Therefore, in this work, we adopt the descriptions generated under the full configuration. \Cref{fig:llm_descriptions} illustrates the initialization process of our LLM-derived classifiers, $W_{\text{LLM}}$ and $W_{\text{LLM}}^*$. Here, $W_{\text{LLM}}$ is initialized with static embeddings, while $W_{\text{LLM}}^*$ is initialized with learnable embeddings. After this initialization step, the CLIP text encoder $E_t$ is discarded and not used during training or inference. In~\cref{tab:gpt3_results}, we incorporate GPT-3 descriptions for the text prototypes in DPA as a fair comparison with \ours and other related SOTA.

However, CuPL employs an older model, GPT-3, to generate its descriptions. As an ablation study, we regenerate descriptions following CuPL’s full configuration using a more recent and capable model, GPT-4o~\cite{achiam2023gpt}. In addition to using CuPL’s original LLM-prompt templates, we design a tailored \textit{system prompt} for GPT-4o: \textcolor{darkbrown}{``You are a helpful assistant. Give 10 numbered sentences answering the prompt as visually identifiable descriptions.''} We further append \textcolor{darkbrown}{``Include `\{CLASS\}' in each sentence.''} to each prompt to ensure consistent mention of the target class in all responses. This level of instruction was unnecessary for GPT-3, which operated as a text completion model rather than a chat-based agent. In~\cref{tab:gpt4o_results}, we compare our method’s performance, using ViT-B/32, with other state-of-the-art (SOTA) approaches. This analysis demonstrates the value of richer and more contextually grounded descriptions in improving \ours. Notably, GPT-4o-generated descriptions lead to higher average accuracy than those from GPT-3. On the fine-grained FGVC Aircraft dataset, our method sees an improvement of up to $24.04\%$, a gain of $1.8\%$ over GPT-3-based descriptions. Similarly, zero-shot accuracy on the Flowers dataset improves by $3.12\%$, contributing to overall performance gains for both \ours and other related methods. DPA remains competitive with \ours on DTD and Flowers, showing gains of $0.06\%$ and $1.00\%$ respectively. We argue that the coarse-grained cues embedded in the pretrained \texttt{[CLS]} token are particularly beneficial for datasets like DTD and Flowers during fine-tuning, as the category of interest in these datasets typically spans the entire image and exhibits spatially diffused features. These limitations are discussed further in \cref{sec:limitations}.

\begin{table}[!ht]
  \centering
  \scriptsize
  \setlength{\tabcolsep}{3pt}
  \renewcommand{\arraystretch}{1.15}
  \begin{tabularx}{\linewidth}{>{\raggedright\arraybackslash}p{2cm} *{7}{>{\centering\arraybackslash}X}}
    \toprule
    \textbf{Component} & 
    \rotatebox{90}{\textbf{Cars}} & 
    \rotatebox{90}{\textbf{DTD}} & 
    \rotatebox{90}{\textbf{FGVC}} & 
    \rotatebox{90}{\textbf{Flowers}} &
    \rotatebox{90}{\textbf{Pets}} & 
    \rotatebox{90}{\textbf{UCF101}} & 
    \rotatebox{90}{\textbf{Avg}} \\
     \midrule
        Zero-shot~\cite{clip} & 60.79 & 50.11 & 20.94 & 69.51 & 61.14 & 66.90 & 54.90 \\
        WCA~\cite{wca} & \underline{61.95} & 51.60 & 21.15 & 68.70 & 86.32 & 65.82 & 59.26 \\
        LaFTer~\cite{lafter} & 57.44 & 50.32 & 19.86 & 72.43 & 84.93 & 65.08 & 58.34 \\
        DPA~\cite{DPA} & 57.32 & \underline{58.60} & \underline{22.08} & \textbf{77.71} & \underline{90.06} & \underline{68.38} & \underline{62.36} \\
        \midrule
    \rowcolor[HTML]{FFD6E7} \ours & \textbf{65.81} & \textbf{60.00} & \textbf{22.74} & \underline{75.84} & \textbf{90.24} & \textbf{70.98} & \textbf{64.27} \\
      \bottomrule
  \end{tabularx}
  \caption{Performance comparison of SOTA methods with GPT-3 generated descriptions.}
\label{tab:gpt3_results}
\end{table}

\begin{table}[!ht]
  \centering
  \scriptsize
  \setlength{\tabcolsep}{3pt}
  \renewcommand{\arraystretch}{1.15}
  \begin{tabularx}{\linewidth}{>{\raggedright\arraybackslash}p{2cm} *{7}{>{\centering\arraybackslash}X}}
    \toprule
    \textbf{Component} & 
    \rotatebox{90}{\textbf{Cars}} & 
    \rotatebox{90}{\textbf{DTD}} & 
    \rotatebox{90}{\textbf{FGVC}} & 
    \rotatebox{90}{\textbf{Flowers}} &
    \rotatebox{90}{\textbf{Pets}} & 
    \rotatebox{90}{\textbf{UCF101}} & 
    \rotatebox{90}{\textbf{Avg}} \\
    \midrule
Zero-shot~\cite{clip} & 58.33 & 52.39 & 21.66 & 72.63 & 88.55 & 65.42 & 59.83 \\
WCA~\cite{wca} & \underline{60.97} & 55.37 & \underline{22.80} & 72.60 & 89.38 & 64.73 & 60.98 \\
LaFTer~\cite{lafter}           & 49.59 & 50.90 & 19.05 & 72.72 & 85.17 & 65.90 & 57.22 \\
DPA~\cite{DPA}              & 57.64 & \textbf{59.26} & 22.23 & \textbf{84.57} & \underline{90.11} & \underline{67.78} & \underline{63.60} \\
\midrule
\rowcolor[HTML]{FFD6E7} \ours      & \textbf{64.30} & \underline{59.20} & \textbf{24.03} & \underline{83.56} & \textbf{90.68} & \textbf{70.00} & \textbf{65.30} \\
\bottomrule
  \end{tabularx}
  \caption{Performance comparison of the SOTA methods using GPT-4o descriptions.}
\label{tab:gpt4o_results}
\end{table}

\subsection{Other Implementation Details}

Unless otherwise specified, we use CLIP~\cite{clip} with a ViT-B/32 backbone for all experiments. Comparisons with RN50 are not feasible for \ours, as our methods and our relevant baselines (e.g., LaFTer~\cite{lafter}, ReCLIP~\cite{reclip}, DPA~\cite{DPA}) are designed specifically for the ViT image encoder architecture.
We apply strong augmentations, including RandomResizedCrop, HorizontalFlip, and RandAugment~\cite{cubuk2020randaugment}, to input images standardized to 224×224 pixels. During training, \ours uses these strong augmentations alongside CenterCrop as the weak augmentation.
For all datasets, we set the learning rate to $10^{-4}$, except for Food101 and SUN397, where it is $10^{-6}$. We employ the AdamW optimizer~\cite{loshchilov2017decoupled} with a cosine learning rate schedule and a batch size of 64 across all datasets and train for 15 epochs. All experiments are conducted on a single NVIDIA A100-SXM4-40GB GPU. While some prior methods (e.g., DPA~\cite{DPA}) tune separate learning rates for each dataset, we find that this approach can lead to overfitting and hinders fair generalization across domains. Instead, inspired by ReCLIP's strategy ~\cite{reclip} on Office-Home—where hyperparameters are selected based on a single domain (Rw)—we tune the learning rate on one representative dataset and apply it uniformly across all benchmarks. This not only promotes consistency and reproducibility but also reduces the risk of dataset-specific over-optimization for both our method and existing baselines~\cite{reclip}.

For crop-based experiments, we follow WCA~\cite{wca} and set the hyperparameters to $\alpha = 0.5$ and $\beta = 0.9$. We use $N = 8$ crops per image, based on the analysis in the main paper, and adopt this value of $N$ for all multi-view alignment experiments (including the training-free WCA), unless otherwise specified. In addition, we set $\gamma = 0.5$, based on the experiments reported in the main paper.

To ensure fair comparisons, we reproduce the results of SOTA methods using their official codebases. We adopt the dataset splits defined by VISSL~\cite{goyal2021vissl} to ensure standardized and reproducible evaluation across benchmarks. 
For ablation experiments in the main paper and supplementary materials, we select 6 of the 13 datasets, following ReCLIP’s procedure~\cite{reclip}. These smaller datasets, chosen for their diverse domains and difficulty levels, enable extensive experimentation while supporting robust generalization evaluation. \cref{tab:detail} provides essential information such as the number of text descriptions per class, the number of classes, and the sizes of both the training and testing sets.

\begin{table}[!ht]
  \centering
  \begin{tabular}{@{}lccrr@{}}
    \toprule
    \textbf{Dataset} & \textbf{Desc/Class} & \textbf{Classes} & \textbf{Train} & \textbf{Test} \\
    \midrule
    Birdsnap        & 30   & 500  & 31,900 & 7,977 \\
    Caltech101      & 30  & 100  & 4,403  & 6,645 \\
    Stanford Cars   & 90  & 196  & 8,144  & 8,041 \\
    CIFAR100        & 40  & 100  & 50,000 & 10,000 \\
    DTD             & 60  & 47   & 3,760  & 1,880 \\
    FGVC            & 20   & 102  & 3,334  & 3,333 \\
    Flowers102      & 20  & 102  & 4,093  & 2,463 \\
    Food101         & 30  & 101  & 75,750 & 25,250 \\
    ImageNet-1K     & 50   & 1000 & 50,000 & 50,000 \\
    Oxford Pets     & 20  & 37   & 3,680  & 3,669 \\
    RESISC45        & 50   & 45   & 25,200 & 6,300 \\
    SUN397          & 30  & 397  & 76,129 & 21,758 \\
    UCF101          & 50  & 101  & 9,537  & 3,783 \\
    \bottomrule
  \end{tabular}
  \caption{Detailed dataset statistics.}
  \label{tab:detail}
\end{table}

\section{Additional Experiments}
\label{sec:addition_exp}
In this section, we present additional experiments to quantitatively analyze the effectiveness of our method. In~\cref{sec:sensitivity_lr}, we perform a sensitivity analysis of \ours with respect to the learning rate. In~\cref{sec:few_shot}, we compare \ours with 1-2 shot methods to demonstrate that it outperforms even these few-shot baselines. 
Finally, in~\cref{sec:other_VLM}, we replace CLIP with MetaCLIP to show that \ours maintains strong performance across different pretrained VLMs.

\subsection{Sensitivity to Learning Rate Selection}
\label{sec:sensitivity_lr}

\begin{figure}[!ht]
    \centering
    \includegraphics[width=0.70\linewidth]{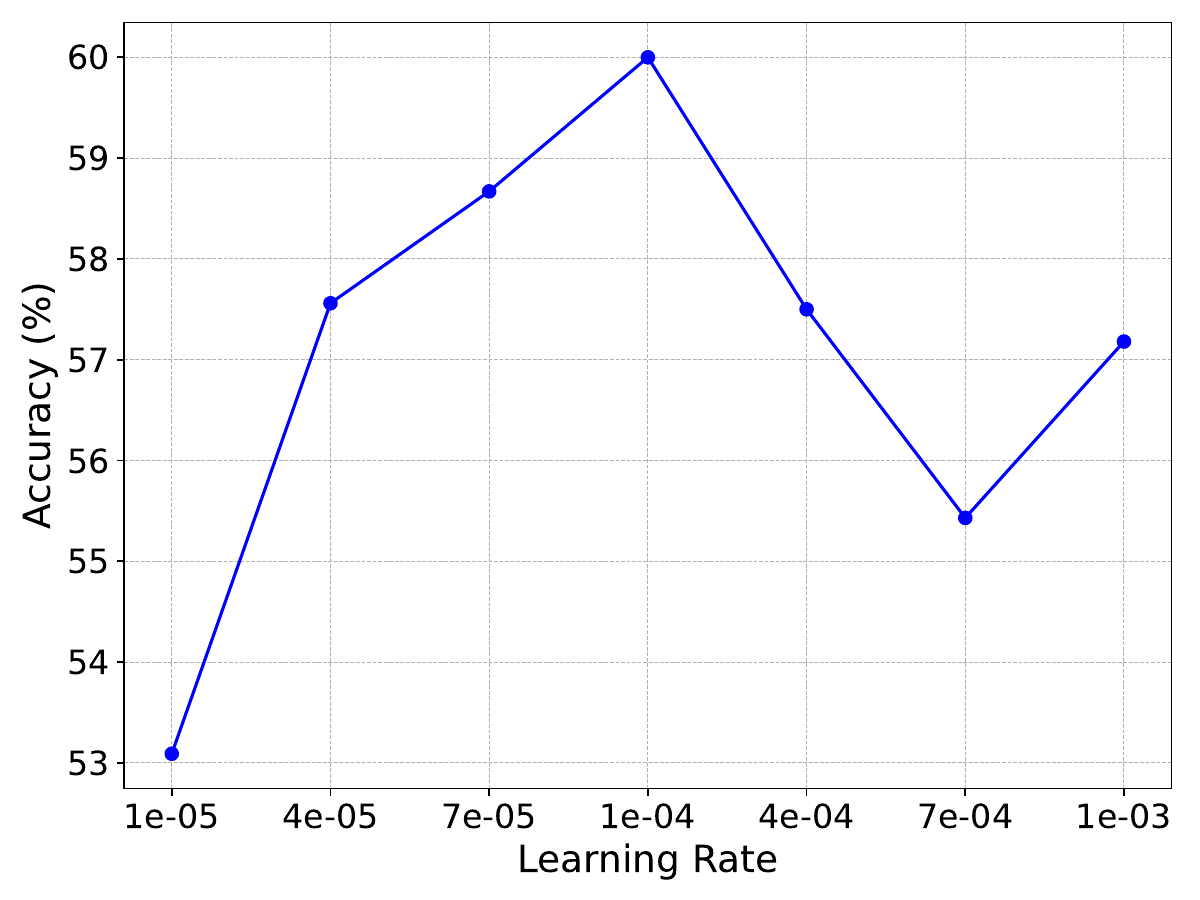}
    \caption{Learning rate selection on the DTD dataset.}
    \label{fig:lr}
\end{figure}

Following ReCLIP~\cite{reclip}, we use a single dataset—DTD~\cite{cimpoi2014describing}—to tune the learning rate and select hyperparameters for both \ours and the SOTA methods. The rationale is to avoid overfitting to any particular test dataset while ensuring a consistent evaluation protocol. The selected hyperparameters are then applied uniformly across all 13 benchmark datasets. As shown in~\cref{fig:lr}, a learning rate of $1\mathrm{e}{-4}$ achieves the highest accuracy on DTD and is chosen as the default. However, for datasets with a large number of classes and greater visual diversity—such as Food101, SUN397, and ImageNet—we reduce the learning rate to $1\mathrm{e}{-6}$ to improve training stability and generalization. To ensure fair comparison, we follow the same tuning procedure and search space for all SOTA.

\subsection{Comparison with Few-Shot Methods}
\label{sec:few_shot}
\Cref{tab:few_shot} compares our method (\ours), which operates in a fully unsupervised setting, against recent few-shot adaptation methods—CoOp~\cite{coop}, MaPLe~\cite{khattakMaPLe}, and CLIP-LoRA~\cite{cliploramaiu}—under 1-shot and 2-shot scenarios. Despite not using any labeled target samples, \ours consistently outperforms all few-shot baselines across most datasets, achieving the highest average accuracy. This demonstrates the robustness and effectiveness of our approach in adapting to target domains without supervision.

\begin{table}[!ht]
  \centering
  \scriptsize
  \setlength{\tabcolsep}{3pt}
  \renewcommand{\arraystretch}{1.15}
  \begin{tabularx}{\linewidth}{>{\raggedright\arraybackslash}p{2cm} *{7}{>{\centering\arraybackslash}X}}
    \toprule
    \textbf{Component} & 
    \rotatebox{90}{\textbf{Cars}} & 
    \rotatebox{90}{\textbf{DTD}} & 
    \rotatebox{90}{\textbf{FGVC}} & 
    \rotatebox{90}{\textbf{Flowers}} &
    \rotatebox{90}{\textbf{Pets}} & 
    \rotatebox{90}{\textbf{UCF101}} & 
    \rotatebox{90}{\textbf{Avg}} \\
    \midrule
    \rowcolor[gray]{0.9} \multicolumn{8}{c}{1-shot} \\
    \midrule
    CoOp~\cite{coop}  & 57.70 & 44.40 & 19.60 & 67.10 & 86.90 & 68.00 & 57.28 \\
    MaPLe~\cite{khattakMaPLe}  & 57.50 & 28.60 & 13.30 & 64.10 & \underline{89.40} & 65.50 & 53.07 \\
    CLIP-LoRA~\cite{cliploramaiu}  & 51.51 & 19.17 & 24.09 & \underline{77.75} & 32.25 & 17.54 & 37.05 \\
    \midrule
    \rowcolor[gray]{0.9} \multicolumn{8}{c}{2-shot} \\
    \midrule
    CoOp   & \underline{62.80} & \underline{48.40} & 22.40 & 75.40 & 88.60 & \textbf{71.40} & \underline{61.50} \\
    MaPLe  & 61.30 & 48.10 & 21.20 & 66.80 & 83.70 & 65.80 & 57.82 \\
    CLIP-LoRA  & 55.12 & 30.61 & \textbf{24.69} & \textbf{84.94} & 49.86 & 34.43 & 46.61 \\
    \midrule
    \rowcolor[HTML]{FFD6E7} \textbf{\ours} (Ours) & \textbf{65.73} & \textbf{59.31} & \underline{22.74} & 75.07 & \textbf{89.56} & \underline{70.82} & \textbf{63.17} \\ 
    \bottomrule
  \end{tabularx}
    \caption{Comparison with few-shot methods across six datasets. All methods are trained and evaluated using the same dataset splits as used in our approach, following the VISSL~\cite{goyal2021vissl} protocol.}
  \label{tab:few_shot}
\end{table}

\subsection{Comparison with other VLMs}
\label{sec:other_VLM}
We evaluate the performance of \ours when applied to the MetaCLIP~\cite{xu2023metaclip} model in~\cref{tab:metaclip_comparison}. MetaCLIP offers two versions of ViT-B/32 models trained on 400M and 2.5B image-text pairs, and we conduct comparisons using both. 
Our method is benchmarked against the zero-shot baseline and the recent strong approach DPA~\cite{DPA}. Across most datasets, \ours consistently improves performance, underscoring its effectiveness in adapting VLMs using fine-grained information. When using the MetaCLIP-400M model, \ours achieves an overall accuracy improvement of $2.24\%$ over DPA. However, DPA slightly outperforms \ours on FGVC, Flowers, and UCF datasets. 
With the larger MetaCLIP-2.5B model, DPA surpasses \ours by a notable margin of $+4.36\%$ on average across those datasets, resulting in a reduced overall accuracy advantage of $0.92\%$ for \ours. This suggests that the well-curated and large-scale pretraining data used in MetaCLIP may lead to better alignment between the visual and textual modalities, thereby enhancing the effectiveness of DPA's dual prototype alignment mechanism. In contrast, \ours relies solely on cues from LLM-generated descriptions to construct its classifier, without explicitly leveraging image prototypes.

\begin{table}[!ht]
  \centering
  \scriptsize
  \setlength{\tabcolsep}{3pt}
  \renewcommand{\arraystretch}{1.15}
  \begin{tabularx}{\linewidth}{>{\raggedright\arraybackslash}p{2.5cm} *{7}{>{\centering\arraybackslash}X}}
    \toprule
    \textbf{Component} & 
    \rotatebox{90}{\textbf{Cars}} & 
    \rotatebox{90}{\textbf{DTD}} & 
    \rotatebox{90}{\textbf{FGVC}} & 
    \rotatebox{90}{\textbf{Flowers}} &
    \rotatebox{90}{\textbf{Pets}} & 
    \rotatebox{90}{\textbf{UCF101}} & 
    \rotatebox{90}{\textbf{Avg}} \\
    \midrule
    \rowcolor[gray]{0.9} \multicolumn{8}{c}{MetaCLIP (ViT-B/32) 400M} \\
    \midrule
    Zero-shot~\cite{clip} & 68.23 & \underline{60.69} & 28.20 & 69.91 & 87.90 & 64.10 & 63.17 \\
    DPA~\cite{DPA}           & \underline{69.40} & 56.90 & \textbf{30.87} & \textbf{76.86} & \underline{89.80} & \textbf{72.10} & \underline{65.99} \\
    \rowcolor[HTML]{FFD6E7} \textbf{\ours} (Ours)
                 & \textbf{74.93} & \textbf{66.60} & \underline{30.12} & \underline{75.52} & \textbf{90.62} & \underline{71.56} & \textbf{68.23} \\
    \midrule
    \rowcolor[gray]{0.9} \multicolumn{8}{c}{MetaCLIP (ViT-B/32) 2.5B} \\
    \midrule
    Zero-shot & 69.60 & 60.96 & 29.79 & 69.47 & 88.50 & 65.40 & 63.95 \\
    DPA           & \underline{76.00} & \underline{61.86} & \underline{30.48} & \underline{75.56} & \textbf{91.50} & \textbf{76.90} & \underline{68.72} \\
    \rowcolor[HTML]{FFD6E7} \textbf{\ours} (Ours)
                 & \textbf{80.66} & \textbf{65.37} & \textbf{31.71} & \textbf{76.82} & \underline{90.73} & \underline{72.54} & \textbf{69.64} \\
    \bottomrule
  \end{tabularx}
  \caption{Performance comparison using MetaCLIP.}
  \label{tab:metaclip_comparison}
\end{table}

\section{Qualitative Analysis} 
\label{sec:qualitative_analysis}

\begin{figure}[!ht]
    \centering
    \includegraphics[width=\linewidth, trim=0 0 80 0, clip]{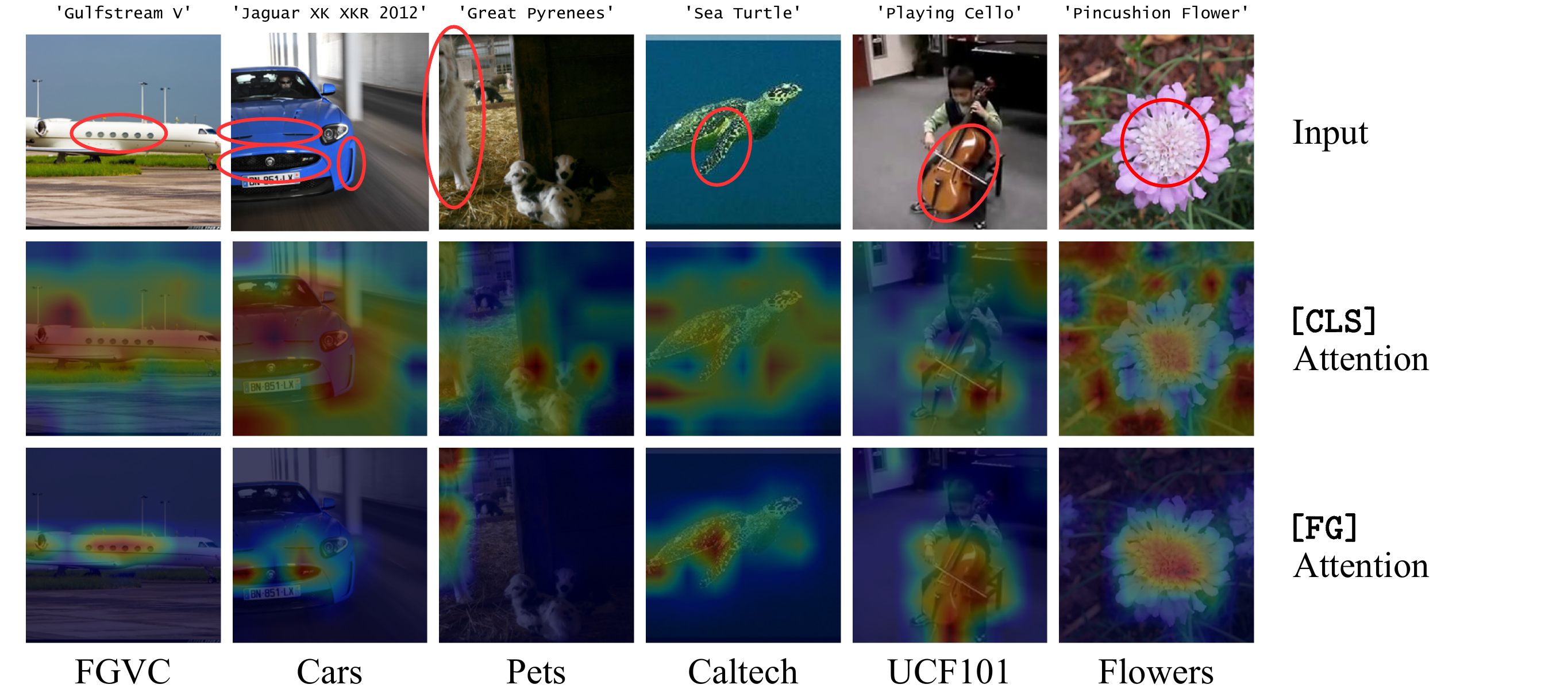}
    \caption{Visualization of attention maps in \ours. \textit{Best viewed zoomed in.}}
    \label{fig:attn_vis}
\end{figure}

In~\cref{fig:attn_vis}, we compare the attention received by patch tokens from the \texttt{[CLS]} token and the \texttt{[FG]} token, showing their distinct focus on global and fine-grained visual patterns, respectively. In the following text, we analyze each image in the figure to highlight how the \texttt{[FG]} token complements the \texttt{[CLS]} token by attending to critical local cues that are essential for fine-grained recognition across different datasets.

\begin{figure}[h]
  \centering
  \includegraphics[width=0.15\textwidth]{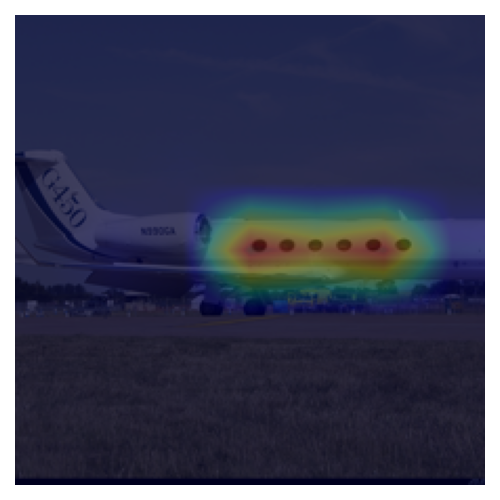}
  \caption{\texttt{`Gulfstream V'}}
  \label{fig:gulfstream_iv}
\end{figure}

~\\
\noindent\textbf{FGVC (Gulfstream V):}
The \texttt{[CLS]} token focuses broadly on the fuselage and wing (including the engine), capturing global shape, but misses the fine-grained details that distinguish the Gulfstream V from similar aircraft. In contrast, the \texttt{[FG]} token accurately attends to the row of six circular windows, a critical cue differentiating the Gulfstream V from the Gulfstream IV (see \cref{fig:gulfstream_iv}), which has only five. This specific localization behavior highlights the \texttt{[FG]} token’s contribution to fine-grained aircraft recognition, complementing the coarse-level attention from \texttt{[CLS]}. This also demonstrates the effectiveness of our design choice to use a learnable classifier initialized with LLM-generated descriptions, which helps align fine-grained visual features with LLM-derived knowledge.

\noindent\textbf{Cars (Jaguar XK XKR 2012):}
While the \texttt{[CLS]} token exhibits broad attention over the hood and front bumper, it neglects finer visual cues. The \texttt{[FG]} token sharply attends to the front grille area, housing the Jaguar logo, as well as the right headlight and hood vent, all of which are discriminative for identifying the Jaguar XKR variant. These details enable the model to correctly resolve both the brand and specific trim level, showcasing how the saliency-oriented \texttt{[FG]} token provides the crucial local context that the pretrained \texttt{[CLS]} token alone cannot capture.

\noindent\textbf{Pets (Great Pyrenees):}
The \texttt{[CLS]} token provides diffuse coverage over the entire scene, including both the sheep and the barn wall, but fails to give the highest focus on the main subject of the image: the dog, which is partially occluded and visually entangled with the background. In contrast, the \texttt{[FG]} token attends sharply to the dog itself, effectively isolating the fine-grained details necessary for accurate identification, thereby correcting the ambiguity introduced by the global attention.

\noindent\textbf{Caltech (Sea Turtle):}
Global attention from the \texttt{[CLS]} token spreads across the body of the sea turtle and the surrounding water, capturing the object in context but without specificity. The \texttt{[FG]} token locks onto the turtle’s textured shell and the flipper (hand) region, critical identifiers for distinguishing a sea turtle from other marine creatures. This focused attention helps refine the representation and improves recognition accuracy by grounding the prediction in discriminative parts.

\noindent\textbf{UCF101 (Playing Cello):}
The \texttt{[CLS]} token's attention broadly spans the person and the background, incorporating contextual cues from the scene such as the instrument and floor. However, the \texttt{[FG]} token mostly focuses on the cello itself, particularly the bow and body, where the action and object interaction occur. This focused attention is crucial for activity recognition, where distinguishing between ``playing cello'' and other musical actions relies on fine-grained spatial relations between the human and the instrument.

\noindent\textbf{Flowers (Pincushion Flower):}
While \texttt{[CLS]} attention distributes itself over the general flower and its surroundings, the \texttt{[FG]} token concentrates precisely on the central cluster of small florets, a key structure that defines the pincushion flower. The fine-grained pattern within the central disk is essential to differentiate this species from visually similar ones.


\section{Normalized Cut Algorithm}
\label{sec:ncut}

In the main paper, we used the $\text{NCut}(v_{\text{patch}})$ notation but omitted its definition. Here, we provide all the mathematical derivations from \cite{TokenCut} for the completeness of our paper.

\subsection{Mathematical Derivation}

Consider a graph $\mathcal{G} = (\mathcal{V}, \mathcal{E})$, with $\mathcal{V}$ as nodes and $\mathcal{E}$ as weighted edges. In our case, all the tokens $v_{\text{patch}}$ are considered as the set of nodes, and the pair-wise affinities between tokens are the set of edges (relations). The main concept behind NCut is graph cuts. Any graph $\mathcal{G}$ can be partitioned into two disjoint sets $\mathcal{A}$ and $\mathcal{B}$, where $\mathcal{A} \cup \mathcal{B}=\mathcal{V}$ and $\mathcal{A} \cap \mathcal{B} = \phi$ by simply removing edges connecting $\mathcal{A}$ and $\mathcal{B}$. In graph theoretic language, the total weight of the edges removed is called the \textit{cut} and it is considered as the degree of dissimilarity between $\mathcal{A}$ and $\mathcal{B}$. This is expressed in \cref{eq:cut}.

\begin{equation}
\label{eq:cut}
    \text{Cut}(\mathcal{A},\mathcal{B})=\sum_{u\in \mathcal{A}, v\in \mathcal{B}}w(u,v)
\end{equation}

\begin{figure*}[!ht]
  \centering
  \begin{subfigure}[b]{0.49\textwidth}
    \centering
    \includegraphics[width=\linewidth]{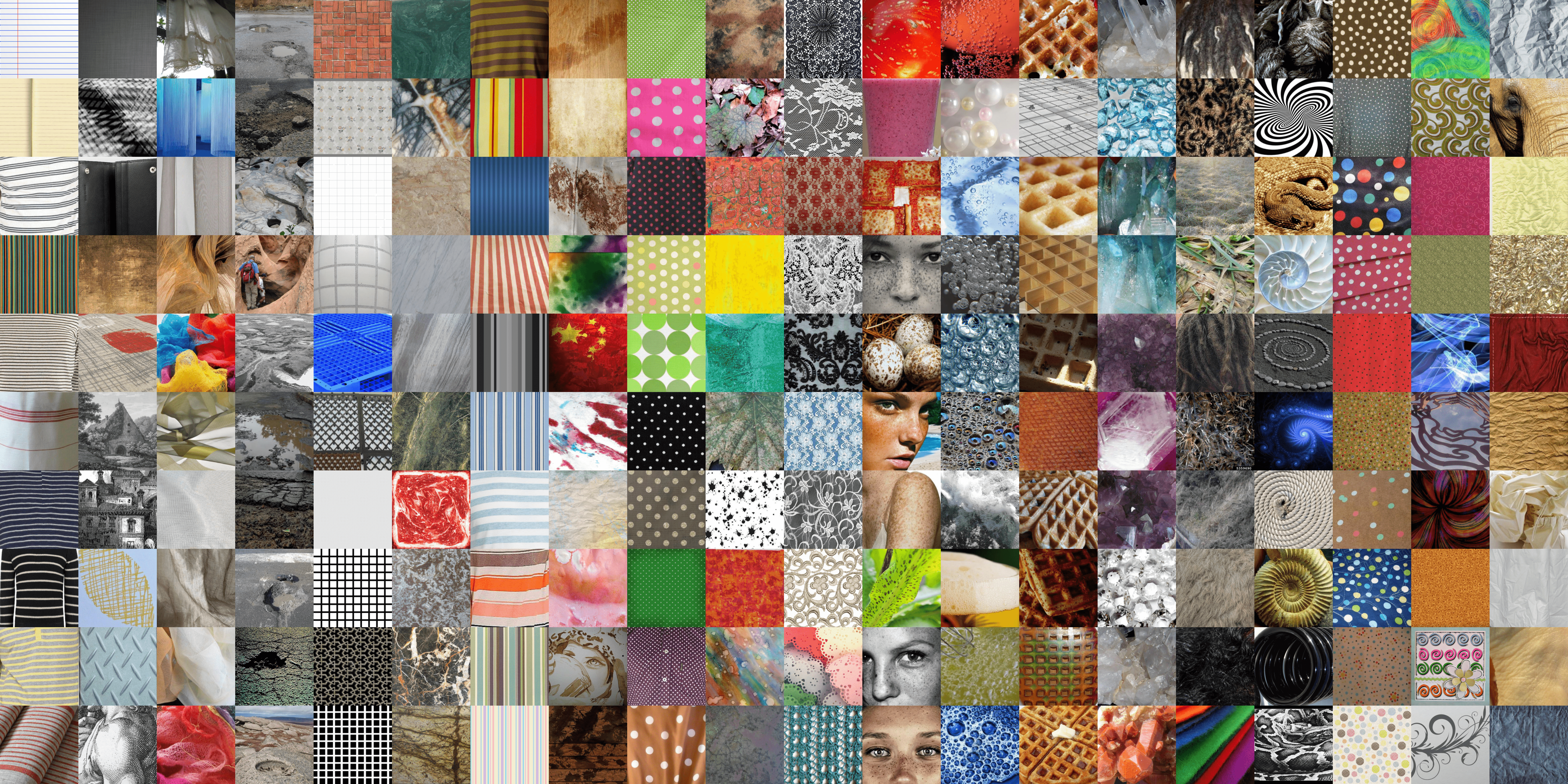}
    \caption{DTD}
    \label{fig:dtd_collage}
  \end{subfigure}\hfill
  \begin{subfigure}[b]{0.49\textwidth}
    \centering
    \includegraphics[width=\linewidth]{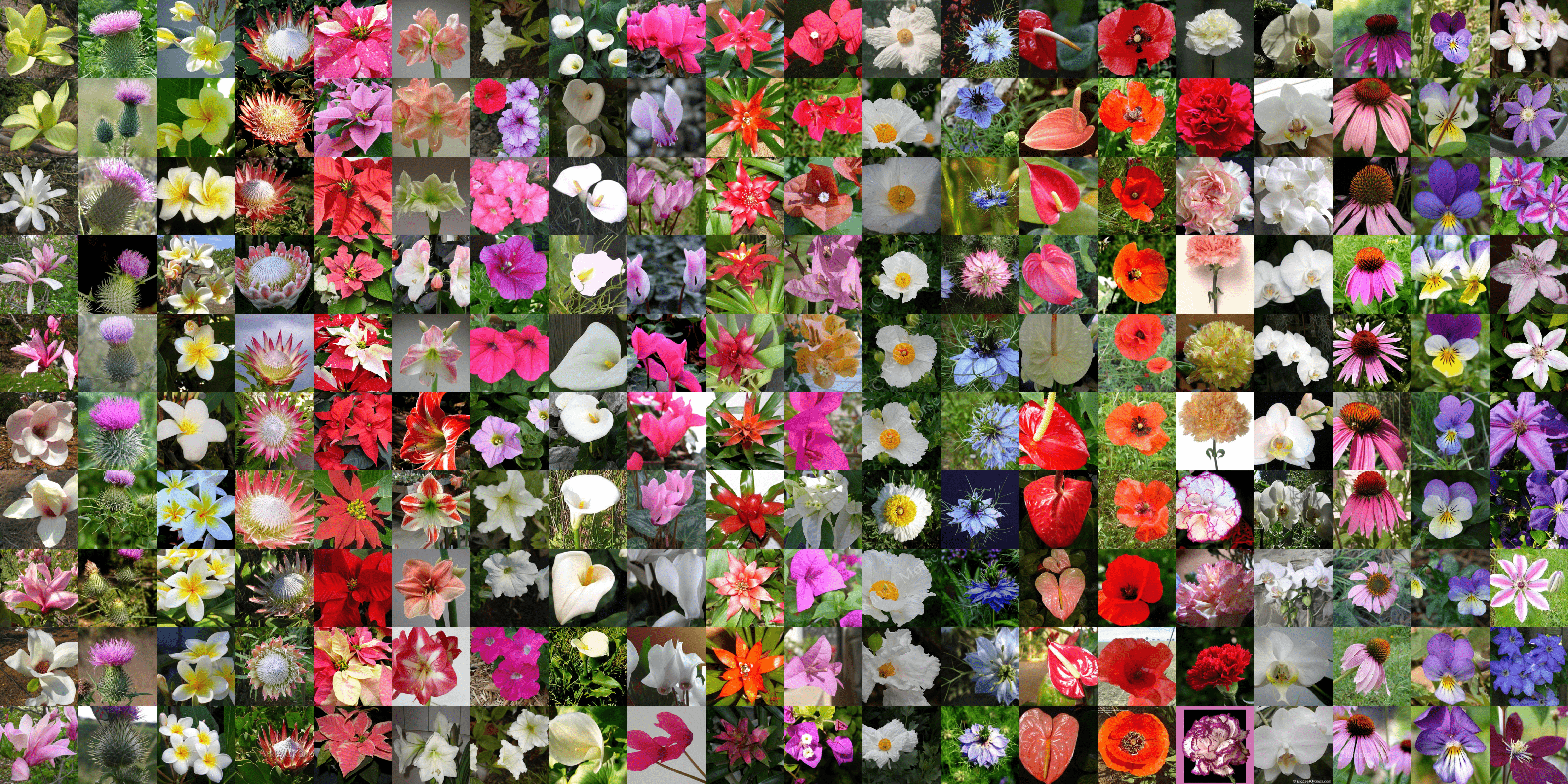}
    \caption{Flowers}
    \label{fig:flowers_collage}
  \end{subfigure}
  \caption{Visual examples from the DTD (a) and Flowers (b) datasets. Categories in both datasets typically span the entire image and exhibit spatially diffuse features without strong localized cues. Such characteristics favor global representations, such as the pretrained \texttt{[CLS]} token.}
  \label{fig:dtd_flowers_subfigs}
\end{figure*}

Let $\mathbf{E}$ be the affinity matrix, where $\mathbf{E}_{i,j}$ represents the edge weight between nodes $v_i$ and $v_j$. The Normalized Cut method~\cite{NCut} computes the optimal cut that partitions $\mathcal{G}$ into disjoint sets $\mathcal{A}$ and $\mathcal{B}$, balancing dissimilarity between sets and similarity within sets. The NCut energy to minimize is:
\begin{equation}
\label{eqn:ncut_energy}
\frac{\operatorname{Cut}(\mathcal{A}, \mathcal{B})}{\operatorname{assoc}(\mathcal{A}, \mathcal{V})} + \frac{\operatorname{Cut}(\mathcal{A}, \mathcal{B})}{\operatorname{assoc}(\mathcal{B}, \mathcal{V})},
\end{equation}
where $\operatorname{assoc}(\mathcal{A}, \mathcal{V})$ is the total similarity from nodes in $\mathcal{A}$ to all nodes. The optimization problem can be reformulated as:
\begin{equation}
\label{eqn:ncut_final}
\min_{\mathbf{y}} \frac{\mathbf{y}^T (\mathbf{D} - \mathbf{E}) \mathbf{y}}{\mathbf{y}^T \mathbf{D} \mathbf{y}},
\end{equation}
subject to $\mathbf{y} \in \{1, -b\}^N$ and $\mathbf{y}^T \mathbf{D} \mathbf{1} = 0$, where $\mathbf{D}$ is a diagonal matrix with $\mathbf{D}_{i,i} = \sum_j \mathbf{E}_{i,j}$.
By setting $\mathbf{z} = \mathbf{D}^{\frac{1}{2}} \mathbf{y}$, the problem becomes:
\begin{equation}
\label{eqn:ncut_z}
\min_{\mathbf{z}} \frac{\mathbf{z}^T \mathbf{D}^{-\frac{1}{2}} (\mathbf{D} - \mathbf{E}) \mathbf{D}^{-\frac{1}{2}} \mathbf{z}}{\mathbf{z}^T \mathbf{z}}.
\end{equation}
This is equivalent to the Rayleigh quotient, corresponding to the eigenvalue problem $\mathbf{D}^{-\frac{1}{2}} (\mathbf{D} - \mathbf{E}) \mathbf{D}^{-\frac{1}{2}} \mathbf{z} = \lambda \mathbf{z}$. Since $\mathbf{D} - \mathbf{E}$ is the positive semidefinite Laplacian matrix, the smallest eigenvalue is $\lambda = 0$ with eigenvector $\mathbf{z}_0 = \mathbf{D}^{\frac{1}{2}} \mathbf{1}$. The second smallest eigenvector $\mathbf{z}_1$, known as Fiedler vector, orthogonal to $\mathbf{z}_0$, minimizes the energy in Eq.~\eqref{eqn:ncut_z}:
\begin{equation}
\mathbf{z}_1 = \underset{\mathbf{z}^T \mathbf{z}_0 = 0}{\text{min}} \frac{\mathbf{z}^T \mathbf{D}^{-\frac{1}{2}} (\mathbf{D} - \mathbf{E}) \mathbf{D}^{-\frac{1}{2}} \mathbf{z}}{\mathbf{z}^T \mathbf{z}}.
\end{equation}

TokenCut~\cite{TokenCut} uses the average value of $z_1$ to cut the patch token graph into most-salient and least-salient regions. The most salient region is filtered out based on the maximum absolute value of the Fiedler vector in the two partitions.

\subsection{Computational Complexity}
\label{sec:ncut_complexity}

Let \(N\) denote the number of patch tokens (typically \(N = P^2\) for a \(P \times P\) grid). Given an affinity matrix \(W \in \mathbb{R}^{N \times N}\), Normalized Cut (NCut) involves solving a spectral partitioning problem via the graph Laplacian \(L = D - W\), where \(D\) is the degree matrix.

\noindent \textbf{Affinity Graph Construction.}  
To make NCut tractable at inference time, we build a sparse affinity graph. This reduces the number of non-zero edges \(|E|\) from \(O(N^2)\) (dense) to \(O(N)\), assuming each node connects to a fixed number of neighbors (e.g., 4- or 8-connected grid). As a result, computing the affinity matrix and degree matrix both take \(O(N)\) time and memory.

\noindent \textbf{Eigenvector Computation.}  
We compute the second smallest eigenvector (Fiedler vector) of the normalized Laplacian using iterative sparse eigensolvers~\cite{Kressner_2023}. These solvers scale as \(O(N)\) per iteration for sparse matrices, and converge in a small number of steps for well-conditioned problems.

\noindent \textbf{Overall Complexity.}  
With sparse affinity and an efficient eigensolver, NCut runs in \textbf{linear time and memory}, i.e., \(O(N)\), making it practical for real-time inference. In contrast, naive dense solvers would require \(O(N^3)\) time and \(O(N^2)\) space, which becomes prohibitive even for modest input sizes.

\noindent \textbf{Practical Feasibility.}  
For a typical CLIP-based vision transformer (consider a relatively big model ViT-B/16, for instance), the number of tokens per image is modest (e.g., \(14 \times 14 = 196\)). Under this setting, NCut runs efficiently and incurs negligible overhead relative to CLIP forward passes.

\section{Limitations and Future Directions}
\label{sec:limitations}

While our approach demonstrates strong performance across a range of tasks, it exhibits certain limitations, particularly in scenarios where a careful balance between local and global information is required during the \ours fine-tuning process. As discussed throughout the main paper, our pseudo-labeler depends entirely on the model being fine-tuned to generate accurate self-labels. This reliance can be limiting when the dataset primarily contains coarse, spatially diffuse features rather than localized, fine-grained cues.
This limitation becomes evident in datasets such as DTD and Flowers, as illustrated in \cref{fig:dtd_flowers_subfigs}. In these datasets, the category of interest typically spans the entire image, and the corresponding features are distributed across the spatial dimensions. In such cases, a symmetric fusion of fine-grained and coarse-grained token predictions may inadvertently introduce localized spatial biases, which are less aligned with the overall structure of the data.
Although our method achieves a $4.04\%$ improvement over DPA on DTD in the main results, DPA remains competitive when equipped with a learnable GPT-3-derived classifier in place of the textual prototypes, with a $1.4\%$ accuracy gap (see \cref{tab:gpt3_results}). Similarly, on the Flowers dataset, DPA performs comparably to \ours in the main results and even surpasses it by $1.87\%$ when using the same GPT-3-derived classifier.
These findings suggest that TokenFusion could benefit from a more flexible fusion strategy, such as an adaptive weighting mechanism between coarse- and fine-grained predictions, rather than relying on a symmetric fusion scheme. We consider this a promising direction for future work.

\section{Pseudocode and Notation}
\label{sec:pseudo_code}

We present the detailed pseudocode of \ours in \cref{alg:TokenFusion_train}.

\begin{algorithm*}[!ht]
    \caption{\ours self-training}\label{alg:TokenFusion_train}
    \begin{algorithmic}[1]
        \Require CLIP vision encoder, $E_v^\Theta$ where $\Theta$ represents all the affine parameters in the LayerNorm layers;
        \Statex \hspace{2.5em} Frozen vision-to-text (shared embedding space) projection function $P_{\text{CLIP}}$;
        \Statex \hspace{2.5em} Learnable attention pooling projection parameters $W_Q, W_K, W_V;$
        \Statex \hspace{2.5em} Unlabeled images of a target dataset $\mathcal{X}_t = \{x_i\}_{i=1}^N$;
        \Statex \hspace{2.5em} An LLM model $h(\cdot)$; Set of class names $\mathcal{Y}$ with $C = |\mathcal{Y}|$;
        \Statex \hspace{2.5em} Multi-crop augmentation $\alpha(\cdot)$; Strong augmentation $\mathcal{A}(\cdot)$;
        \Statex \hspace{2.5em} Cosine similarity function $s(\cdot, \cdot)$;
        \Statex \hspace{2.5em} Knowledge weighting coefficient $\gamma$;
        \Statex \hspace{2.5em} Number of epochs $\texttt{MaxEpochs}$; Batch size $\texttt{B}$
        \Function{InitClassifiers}{$E_t$, $\mathcal{Y}$, h}
            \State $\mathbf{W} \gets \{\emptyset\}_{j=1}^C$
            \ForEach{$y \in \mathcal{Y}$}
                \State $\mathbf{t} \gets h(y)$ \Comment{Prompt the LLM to extract $M$ number of descriptions for class $y$}
                \State $\mathbf{W}_j \gets \frac{1}{M}\sum_{i=1}^M E_t(\mathbf{t})$ \Comment{Average of the description embedding $W_j \in \mathbb{R}^{1\times d}$ for class $y$}
            \EndFor
            \State \Return NoBackProp$(\mathbf{W})$, $\mathbf{W}$
        \EndFunction
        ~\\
        \State $W_{\text{LLM}}$, $W_{\text{LLM}}^{*} \gets$ \Call{InitClassifiers}{$E_t$, $\mathcal{Y}$, h} \Comment{Initialize $W_{\text{LLM}}$ and $W_{\text{LLM}}^* \in \mathbb{R}^{C\times d}$}
        ~\\
        \Function{$f_{\text{AttnPool}}$}{$q$, $v_{\text{patch}}$}
            \State $\text{scores} \gets \text{softmax}\left( \frac{q_{\text{sal}} W_{\text{Q}} (v_{\text{patch}} W_{\text{K}})^\top}{\sqrt{d}} \right)$
            \State \Return $\text{scores} \cdot (v_{\text{patch}} W_{\text{V}})$
        \EndFunction
        ~\\
        \Function{\text{TokenFusion}}{$x$, $W_{\text{LLM}}^*$} \Comment{$x\in \mathbb{R}^{1\times W\times H\times 3}$}
            \State $[v_{\text{patch}}, v^{\texttt{CLS}}] \gets E_v(x)$ \Comment{$v_{\text{patch}} \in \mathbb{R}^{n\times d}, v^{\texttt{CLS}}\in \mathbb{R}^d$}
            \State $ \mathcal{V}_{\text{cut}} \gets \text{NCut}(v_{\text{patch}})$ \Comment{Apply Normalized Cut algorithm}
            \State $q_{\text{sal}} \gets \frac{1}{|\mathcal{V}_{cut}|}\sum_{\forall v \in \mathcal{V}_{cut}}\!\!\!\!v$ \Comment{Creation of the saliency-oriented query}
            \State $v^{\texttt{FG}} \gets f_{\text{AttnPool}}(q_{\text{sal}}, v_{\text{patch}})$ \Comment{Attention pooling with $q_{\text{sal}}$ to form the fine-grained token}
            \State $\text{Logits}_{\text{local}} \gets s(P_{\text{CLIP}}(v^{\texttt{FG}}) , W_{\text{LLM}}^*)$ \Comment{Logits from the fine-grained token}
            \State $\text{Logits}_{\text{global}} \gets s(P_{\text{CLIP}}(v^{\texttt{CLS}}), W_{\text{LLM}}^*)$ \Comment{Logits from the coarse-grained token}
            \State \Return $\frac{\text{Logits}_{\text{local}} + \text{Logits}_{\text{global}}}{2}$ \Comment{Symmetric fusion of the logits}
        \EndFunction
        ~\\
        \Function{MultiViewAlignment}{$x$, $W_{\text{LLM}}$}
            \State $[\_, v^{\texttt{CLS}}] \gets E_v(x)$
            \State $f(x) \gets P_{\text{CLIP}}(v^{\texttt{CLS}})$
            \State $\alpha(x) \gets \left\{x_i | x_i = \phi(x, \lambda_i \cdot \min(H, W)) \mid i = 1, \dots, N \right\}$
            \State $w_i \gets \frac{\exp\left(s(f(x), f(x_i))\right)}{\sum_{l=1}^{N} \exp\left(s(f(x), f(x_l))\right)}$
            \State $f^{\text{agg}}(x) \gets \sum_{i=1}^N w_i \cdot f(x_i | \alpha)$
            \State \Return $s(f^{\text{agg}}(x), W_{\text{LLM}})$
        \EndFunction
        ~\\
        \For{$\texttt{epoch} \gets 1$ to $\texttt{MaxEpochs}$}
            \State $\mathbf{x} \gets$ \Call{SampleMiniBatch}{$\mathcal{X}_t$, $B$} \Comment{$\mathbf{x} \in \mathbb{R}^{B\times W\times H\times 3}$}
            ~\\
            \State With \textbf{no Back-Propagation}:
                \State \hspace{1.5em} $\text{Pseudo-logits}_{\text{CLIP}} \gets$ \Call{MultiViewAlignment}{$\mathbf{x}$, $W_{\text{LLM}}$}
                \State \hspace{1.5em} $\hat{y} \gets \arg\max_{y \in \mathcal{Y}} \{\gamma \cdot \text{Pseudo-logits}_{\text{CLIP}} + (1-\gamma) \cdot \text{TokenFusion}(\mathbf{x}, W_{\text{LLM}}^*)\}$ \Comment{Dynamic Knowledge Aggregation}
            ~\\
            \State $p_{\mathcal{A}(x)} \gets \text{softmax}($\Call{TokenFusion}{$\mathcal{A}(\mathbf{x})$, $W_{\text{LLM}}^*$} $, axis=1)$ \Comment{Strongly-augmented counterpart}
            \State $\mathcal{L}_{st} \gets \text{CrossEntropy}(p_{\mathcal{A}(x)}, \hat{y})$ \Comment{Self-training loss}
            \State $\mathcal{L}_{reg} \gets -\frac{1}{C} \sum_{j=1}^C \log\left(\bar{p}_{\mathcal{A}(x),j}\right)$ \Comment{Fairness regularization loss}
            \State $\mathcal{L} \gets \mathcal{L}_{st} + \mathcal{L}_{reg}$
            \State \textbf{Back-Propagate} over $\Theta$, $W_{\text{LLM}}^{*}$, $W_Q$, $W_K$ and $W_V$ on $\mathcal{L}$
        \EndFor
    \end{algorithmic}
\end{algorithm*}

\begin{table*}[!ht]
\centering
\begin{tabularx}{\textwidth}{>{\raggedright\arraybackslash}p{0.2\textwidth}  
                                >{\raggedright\arraybackslash}X}            
\toprule
\textbf{Symbol} & \textbf{Description} \\
\midrule
\rowcolor[gray]{0.9} \multicolumn{2}{c}{\textbf{List of Abbreviations}} \\
\midrule
VLMs & Vision Language Models\\
LLMs & Large Language Models\\
CLIP & Contrastive Language-Image Pretraining\\
UA & Unsupervised Adaptation\\
SOAP & Saliency-Oriented Attention Pooling\\
\midrule
\rowcolor[gray]{0.9} \multicolumn{2}{c}{\textbf{List of Symbols}} \\
\midrule
$E_v$ & The visual encoder of CLIP\\
$E_t$ & The natural language encoder of CLIP\\
$\mathcal{D}_s$ & The source dataset for pre-training CLIP\\
$\mathcal{D}_t$ & The target dataset\\
$\mathcal{X}_t$ & The set of unlabeled images in the target dataset\\
$x$ & An arbitrary unlabeled image sampled from the set of unlabeled images in the target dataset\\
$\mathcal{Y}$ & The set of unique class names\\
$C$ & The number of classes in \(\mathcal{D}_t\)  \\
$y$ & The class name corresponding to the image $x$, sampled from the set of class names\\
$W_{\text{LLM}}$ & Frozen LLM-derived classifier embeddings used for multi-view alignment in the pseudo-labeler\\
$W_{\text{LLM}}^*$ & Learnable LLM-derived classifier embeddings used in TokenFusion module\\
$\text{NCut}(\cdot)$ & Normalized Cut Algorithm\\
$x_{\text{patch}}$ & Sequence of $N$ patch tokens returned at the last layer of $E_v$\\
$v_{\text{patch}}$ & Last layer attention-bypassed patch tokens returned at the penultimate layer of $E_v$\\
$\tilde{x}^{L-1}_{\text{patch}}$ & Patch tokens from the penultimate layer of $E_V$ before passing the last layer\\
$\widetilde{W}_V^L$ & Attention value projection of the last layer of $E_v$\\
$\tilde{v}_{\text{patch}}$ & Last layer attention-bypassed patch tokens before passing through last layer MLP\\
$\mathcal{V}_{\text{cut}}$ & Subset of patch tokens selected by the NCut algorithm\\
$v^{\texttt{CLS}}$ & CLS (global) token\\
$v^{\texttt{FG}}$ & FG (fine-grained) token\\
$q_{\text{sal}}$ & Saliency-oriented query for SOAP\\
$W_Q$ & Query projection for SOAP\\
$W_K$ & Key projection for SOAP\\
$W_V$ & Value projection for SOAP\\
$d$ & Embedding dimensionality of the vision encoder $E_v$\\
$\gamma$ & Knowledge weighting coefficient\\
$P_{\text{CLIP}}$ & Vision-to-text projection\\
$\mathcal{L}_{\text{st}}$ & The self-training loss function \\
$\mathcal{L}_{\text{reg}}$ & The fairness regularization loss function\\
$\alpha(\cdot)$ & Multi-crop augmentation function\\
$\mathcal{A}(\cdot)$  & The strongly-augmented function \\
$\bar{p}_{\mathcal{A}(x)}$ & The model's average prediction from the strongly augmented images across the batch \\

\bottomrule
\end{tabularx}
\end{table*}

{
    \small
    \bibliographystyle{ieee_fullname}
    \bibliography{main}

\begin{thebibliography}{10}\itemsep=-1pt

\bibitem{achiam2023gpt}
Josh Achiam, Steven Adler, Sandhini Agarwal, Lama Ahmad, Ilge Akkaya, Florencia~Leoni Aleman, Diogo Almeida, Janko Altenschmidt, Sam Altman, Shyamal Anadkat, et~al.
\newblock Gpt-4 technical report.
\newblock {\em arXiv preprint arXiv:2303.08774}, 2023.

\bibitem{DPA}
Eman Ali, Sathira Silva, and Muhammad~Haris Khan.
\newblock Dpa: Dual prototypes alignment for unsupervised adaptation of vision-language models.
\newblock In {\em 2025 IEEE/CVF Winter Conference on Applications of Computer Vision (WACV)}, pages 6083--6093. IEEE, 2025.

\bibitem{ba2016layer}
Jimmy~Lei Ba, Jamie~Ryan Kiros, and Geoffrey~E Hinton.
\newblock Layer normalization.
\newblock {\em arXiv preprint arXiv:1607.06450}, 2016.

\bibitem{birdsnap}
Thomas Berg, Jiongxin Liu, Seung Woo~Lee, Michelle~L Alexander, David~W Jacobs, and Peter~N Belhumeur.
\newblock Birdsnap: Large-scale fine-grained visual categorization of birds.
\newblock In {\em Proceedings of the IEEE conference on computer vision and pattern recognition}, pages 2011--2018, 2014.

\bibitem{borji2019salient}
Ali Borji, Ming-Ming Cheng, Qibin Hou, Huaizu Jiang, and Jia Li.
\newblock Salient object detection: A survey.
\newblock {\em Computational visual media}, 5:117--150, 2019.

\bibitem{bossard2014food}
Lukas Bossard, Matthieu Guillaumin, and Luc Van~Gool.
\newblock Food-101--mining discriminative components with random forests.
\newblock In {\em European conference on computer vision}, pages 446--461. Springer, 2014.

\bibitem{Cheng2017RemoteSI}
Gong Cheng, Junwei Han, and Xiaoqiang Lu.
\newblock Remote sensing image scene classification: Benchmark and state of the art.
\newblock {\em Proceedings of the IEEE}, 105(10):1865--1883, 2017.

\bibitem{cimpoi2014describing}
Mircea Cimpoi, Subhransu Maji, Iasonas Kokkinos, Sammy Mohamed, and Andrea Vedaldi.
\newblock Describing textures in the wild.
\newblock In {\em Proceedings of the IEEE conference on computer vision and pattern recognition}, pages 3606--3613, 2014.

\bibitem{cubuk2020randaugment}
Ekin~D Cubuk, Barret Zoph, Jonathon Shlens, and Quoc~V Le.
\newblock Randaugment: Practical automated data augmentation with a reduced search space.
\newblock In {\em Proceedings of the IEEE/CVF conference on computer vision and pattern recognition workshops}, pages 702--703, 2020.

\bibitem{deng2009imagenet}
Jia Deng, Wei Dong, Richard Socher, Li-Jia Li, Kai Li, and Li Fei-Fei.
\newblock Imagenet: A large-scale hierarchical image database.
\newblock In {\em 2009 IEEE conference on computer vision and pattern recognition}, pages 248--255. IEEE, 2009.

\bibitem{fan2024improving}
Lijie Fan, Dilip Krishnan, Phillip Isola, Dina Katabi, and Yonglong Tian.
\newblock Improving clip training with language rewrites.
\newblock {\em Advances in Neural Information Processing Systems}, 36:35544--35575, 2023.

\bibitem{fei2004learning}
Li Fei-Fei, Rob Fergus, and Pietro Perona.
\newblock Learning generative visual models from few training examples: An incremental bayesian approach tested on 101 object categories.
\newblock In {\em 2004 conference on computer vision and pattern recognition workshop}, pages 178--178. IEEE, 2004.

\bibitem{goyal2021vissl}
Priya Goyal, Quentin Duval, Jeremy Reizenstein, Matthew Leavitt, Min Xu, Benjamin Lefaudeux, Mannat Singh, Vinicius Reis, Mathilde Caron, Piotr Bojanowski, Armand Joulin, and Ishan Misra.
\newblock Vissl.
\newblock \url{https://github.com/facebookresearch/vissl}, 2021.

\bibitem{reclip}
Xuefeng Hu, Ke Zhang, Lu Xia, Albert Chen, Jiajia Luo, Yuyin Sun, Ken Wang, Nan Qiao, Xiao Zeng, Min Sun, et~al.
\newblock Reclip: Refine contrastive language image pre-training with source free domain adaptation.
\newblock In {\em Proceedings of the IEEE/CVF Winter Conference on Applications of Computer Vision}, pages 2994--3003, 2024.

\bibitem{upl}
Tony Huang, Jack Chu, and Fangyun Wei.
\newblock Unsupervised prompt learning for vision-language models.
\newblock {\em arXiv preprint arXiv:2204.03649}, 2022.

\bibitem{align}
Chao Jia, Yinfei Yang, Ye Xia, Yi-Ting Chen, Zarana Parekh, Hieu Pham, Quoc Le, Yun-Hsuan Sung, Zhen Li, and Tom Duerig.
\newblock Scaling up visual and vision-language representation learning with noisy text supervision.
\newblock In {\em International conference on machine learning}, pages 4904--4916. PMLR, 2021.

\bibitem{khattakMaPLe}
Muhammad~Uzair Khattak, Hanoona Rasheed, Muhammad Maaz, Salman Khan, and Fahad~Shahbaz Khan.
\newblock Maple: Multi-modal prompt learning.
\newblock In {\em Proceedings of the IEEE/CVF conference on computer vision and pattern recognition}, pages 19113--19122, 2023.

\bibitem{kim2025cosmos}
Sanghwan Kim, Rui Xiao, Mariana-Iuliana Georgescu, Stephan Alaniz, and Zeynep Akata.
\newblock Cosmos: Cross-modality self-distillation for vision language pre-training.
\newblock In {\em Proceedings of the Computer Vision and Pattern Recognition Conference}, pages 14690--14700, 2025.

\bibitem{krause20133d}
Jonathan Krause, Michael Stark, Jia Deng, and Li Fei-Fei.
\newblock 3d object representations for fine-grained categorization.
\newblock In {\em Proceedings of the IEEE international conference on computer vision workshops}, pages 554--561, 2013.

\bibitem{Kressner_2023}
Daniel Kressner, Yuxin Ma, and Meiyue Shao.
\newblock A mixed precision lobpcg algorithm.
\newblock {\em Numerical Algorithms}, 94(4):1653–1671, May 2023.

\bibitem{krizhevsky2009learning}
Alex Krizhevsky and Geoffrey Hinton.
\newblock Learning multiple layers of features from tiny images.
\newblock Technical Report~0, University of Toronto, Toronto, Ontario, 2009.

\bibitem{lafon2024gallop}
Marc Lafon, Elias Ramzi, Cl{\'e}ment Rambour, Nicolas Audebert, and Nicolas Thome.
\newblock Gallop: Learning global and local prompts for vision-language models.
\newblock In {\em European Conference on Computer Vision}, pages 264--282. Springer, 2024.

\bibitem{BLIP}
Junnan Li, Dongxu Li, Caiming Xiong, and Steven Hoi.
\newblock Blip: Bootstrapping language-image pre-training for unified vision-language understanding and generation.
\newblock In {\em International conference on machine learning}, pages 12888--12900. PMLR, 2022.

\bibitem{wca}
Jinhao Li, Haopeng Li, Sarah Erfani, Lei Feng, James Bailey, and Feng Liu.
\newblock Visual-text cross alignment: Refining the similarity score in vision-language models.
\newblock In {\em International Conference on Machine Learning}, 2024.

\bibitem{li2022masked}
Junnan Li, Silvio Savarese, and Steven~CH Hoi.
\newblock Masked unsupervised self-training for label-free image classification.
\newblock {\em International Conference on Learning Representations}, 2023.

\bibitem{ALBEF}
Junnan Li, Ramprasaath Selvaraju, Akhilesh Gotmare, Shafiq Joty, Caiming Xiong, and Steven Chu~Hong Hoi.
\newblock Align before fuse: Vision and language representation learning with momentum distillation.
\newblock {\em Advances in neural information processing systems}, 34:9694--9705, 2021.

\bibitem{li2022supervision}
Yangguang Li, Feng Liang, Lichen Zhao, Yufeng Cui, Wanli Ouyang, Jing Shao, Fengwei Yu, and Junjie Yan.
\newblock Supervision exists everywhere: A data efficient contrastive language-image pre-training paradigm.
\newblock {\em International Conference on Learning Representations}, 2022.

\bibitem{lin2024tagclip}
Yuqi Lin, Minghao Chen, Kaipeng Zhang, Hengjia Li, Mingming Li, Zheng Yang, Dongqin Lv, Binbin Lin, Haifeng Liu, and Deng Cai.
\newblock Tagclip: A local-to-global framework to enhance open-vocabulary multi-label classification of clip without training.
\newblock In {\em Proceedings of the AAAI Conference on Artificial Intelligence}, volume~38, pages 3513--3521, 2024.

\bibitem{liu2010learning}
Tie Liu, Zejian Yuan, Jian Sun, Jingdong Wang, Nanning Zheng, Xiaoou Tang, and Heung-Yeung Shum.
\newblock Learning to detect a salient object.
\newblock {\em IEEE Transactions on Pattern Analysis and Machine Intelligence}, 33(2):353--367, 2010.

\bibitem{loshchilov2017decoupled}
Ilya Loshchilov and Frank Hutter.
\newblock Decoupled weight decay regularization.
\newblock In {\em International Conference on Learning Representations}, 2017.

\bibitem{lu2024rethinkingvisualcontentrefinement}
Jinda Lu, Shuo Wang, Yanbin Hao, Haifeng Liu, Xiang Wang, and Meng Wang.
\newblock Rethinking visual content refinement in low-shot clip adaptation.
\newblock {\em arXiv preprint arXiv:2407.14117}, 2024.

\bibitem{maji2013fine}
Subhransu Maji, Esa Rahtu, Juho Kannala, Matthew Blaschko, and Andrea Vedaldi.
\newblock Fine-grained visual classification of aircraft.
\newblock {\em arXiv preprint arXiv:1306.5151}, 2013.

\bibitem{lafter}
Muhammad~Jehanzeb Mirza, Leonid Karlinsky, Wei Lin, Horst Possegger, Mateusz Kozinski, Rogerio Feris, and Horst Bischof.
\newblock Lafter: Label-free tuning of zero-shot classifier using language and unlabeled image collections.
\newblock {\em Advances in Neural Information Processing Systems}, 36:5765--5777, 2023.

\bibitem{detailclip}
Amin~Karimi Monsefi, Kishore~Prakash Sailaja, Ali Alilooee, Ser-Nam Lim, and Rajiv Ramnath.
\newblock Detailclip: Detail-oriented clip for fine-grained tasks.
\newblock {\em arXiv preprint arXiv:2409.06809}, 2024.

\bibitem{nilsback2008automated}
Maria-Elena Nilsback and Andrew Zisserman.
\newblock Automated flower classification over a large number of classes.
\newblock In {\em 2008 Sixth Indian conference on computer vision, graphics \& image processing}, pages 722--729. IEEE, 2008.

\bibitem{parkhi2012cats}
Omkar~M. Parkhi, Andrea Vedaldi, Andrew Zisserman, and C.~V. Jawahar.
\newblock Cats and dogs.
\newblock {\em IEEE Conference on Computer Vision and Pattern Recognition}, pages 3498--3505, 2012.

\bibitem{perazzi2012saliency}
Federico Perazzi, Philipp Kr{\"a}henb{\"u}hl, Yael Pritch, and Alexander Hornung.
\newblock Saliency filters: Contrast based filtering for salient region detection.
\newblock In {\em 2012 IEEE conference on Computer Vision and Pattern Recognition}, pages 733--740. IEEE, 2012.

\bibitem{CuPL}
Sarah Pratt, Ian Covert, Rosanne Liu, and Ali Farhadi.
\newblock What does a platypus look like? generating customized prompts for zero-shot image classification.
\newblock In {\em Proceedings of the IEEE/CVF International Conference on Computer Vision}, pages 15691--15701, 2023.

\bibitem{clip}
Alec Radford, Jong~Wook Kim, Chris Hallacy, Aditya Ramesh, Gabriel Goh, Sandhini Agarwal, Girish Sastry, Amanda Askell, Pamela Mishkin, Jack Clark, et~al.
\newblock Learning transferable visual models from natural language supervision.
\newblock In {\em International conference on machine learning}, pages 8748--8763. PMLR, 2021.

\bibitem{ranasinghe2023perceptual}
Kanchana Ranasinghe, Brandon McKinzie, Sachin Ravi, Yinfei Yang, Alexander Toshev, and Jonathon Shlens.
\newblock Perceptual grouping in contrastive vision-language models.
\newblock In {\em Proceedings of the IEEE/CVF International Conference on Computer Vision}, pages 5571--5584, 2023.

\bibitem{GradCAM}
Ramprasaath~R Selvaraju, Michael Cogswell, Abhishek Das, Ramakrishna Vedantam, Devi Parikh, and Dhruv Batra.
\newblock Grad-cam: Visual explanations from deep networks via gradient-based localization.
\newblock In {\em Proceedings of the IEEE international conference on computer vision}, pages 618--626, 2017.

\bibitem{NCut}
Jianbo Shi and Jitendra Malik.
\newblock Normalized cuts and image segmentation.
\newblock {\em IEEE Transactions on Pattern Analysis and Machine Intelligence}, 22(8):888--905, 2000.

\bibitem{Shin2022UnsupervisedSO}
Gyungin Shin, Samuel Albanie, and Weidi Xie.
\newblock Unsupervised salient object detection with spectral cluster voting.
\newblock In {\em Proceedings of the IEEE/CVF Conference on Computer Vision and Pattern Recognition}, pages 3971--3980, 2022.

\bibitem{simeoni2023unsupervised}
Oriane Sim{\'e}oni, Chlo{\'e} Sekkat, Gilles Puy, Anton{\'\i}n Vobeck{\`y}, {\'E}loi Zablocki, and Patrick P{\'e}rez.
\newblock Unsupervised object localization: Observing the background to discover objects.
\newblock In {\em Proceedings of the IEEE/CVF Conference on Computer Vision and Pattern Recognition}, pages 3176--3186, 2023.

\bibitem{FLAVA}
Amanpreet Singh, Ronghang Hu, Vedanuj Goswami, Guillaume Couairon, Wojciech Galuba, Marcus Rohrbach, and Douwe Kiela.
\newblock Flava: A foundational language and vision alignment model.
\newblock In {\em Proceedings of the IEEE/CVF conference on computer vision and pattern recognition}, pages 15638--15650, 2022.

\bibitem{soomro2012ucf101}
Khurram Soomro, Amir~Roshan Zamir, and Mubarak Shah.
\newblock Ucf101: A dataset of 101 human actions classes from videos in the wild.
\newblock {\em Center for Research in Computer Vision}, 2012.

\bibitem{pouf}
Korawat Tanwisuth, Shujian Zhang, Huangjie Zheng, Pengcheng He, and Mingyuan Zhou.
\newblock Pouf: Prompt-oriented unsupervised fine-tuning for large pre-trained models.
\newblock In {\em International Conference on Machine Learning}, pages 33816--33832. PMLR, 2023.

\bibitem{wang2020tent}
Dequan Wang, Evan Shelhamer, Shaoteng Liu, Bruno Olshausen, and Trevor Darrell.
\newblock Tent: Fully test-time adaptation by entropy minimization.
\newblock In {\em International Conference on Learning Representations}, 2021.

\bibitem{wang2024sclip}
Feng Wang, Jieru Mei, and Alan Yuille.
\newblock Sclip: Rethinking self-attention for dense vision-language inference.
\newblock In {\em European Conference on Computer Vision}, pages 315--332. Springer, 2024.

\bibitem{Wang2022DebiasedLF}
Xudong Wang, Zhirong Wu, Long Lian, and Stella~X Yu.
\newblock Debiased learning from naturally imbalanced pseudo-labels.
\newblock In {\em Proceedings of the IEEE/CVF Conference on Computer Vision and Pattern Recognition}, pages 14647--14657, 2022.

\bibitem{TokenCut}
Yangtao Wang, Xi Shen, Yuan Yuan, Yuming Du, Maomao Li, Shell~Xu Hu, James~L Crowley, and Dominique Vaufreydaz.
\newblock Tokencut: Segmenting objects in images and videos with self-supervised transformer and normalized cut.
\newblock {\em IEEE Transactions on Pattern Analysis and Machine Intelligence}, 45(12):15790--15801, 2023.

\bibitem{wanyan2023dinomc}
Xinye Wanyan, Sachith Seneviratne, Shuchang Shen, and Michael Kirley.
\newblock Dino-mc: Self-supervised contrastive learning for remote sensing imagery with multi-sized local crops.
\newblock {\em arXiv preprint arXiv:2303.06670}, 2(6):26, 2023.

\bibitem{xiao2010sun}
Jianxiong Xiao, James Hays, Krista~A Ehinger, Aude Oliva, and Antonio Torralba.
\newblock Sun database: Large-scale scene recognition from abbey to zoo.
\newblock In {\em 2010 IEEE computer society conference on computer vision and pattern recognition}, pages 3485--3492. IEEE, 2010.

\bibitem{xiao2024flair}
Rui Xiao, Sanghwan Kim, Mariana-Iuliana Georgescu, Zeynep Akata, and Stephan Alaniz.
\newblock Flair: Vlm with fine-grained language-informed image representations.
\newblock In {\em Proceedings of the Computer Vision and Pattern Recognition Conference}, pages 24884--24894, 2025.

\bibitem{xu2023metaclip}
Hu Xu, Saining Xie, Xiaoqing~Ellen Tan, Po-Yao Huang, Russell Howes, Vasu Sharma, Shang-Wen Li, Gargi Ghosh, Luke Zettlemoyer, and Christoph Feichtenhofer.
\newblock Demystifying clip data.
\newblock {\em arXiv preprint arXiv:2309.16671}, 2023.

\bibitem{xu2023demystifying}
Hu Xu, Saining Xie, Xiaoqing~Ellen Tan, Po-Yao Huang, Russell Howes, Vasu Sharma, Shang-Wen Li, Gargi Ghosh, Luke Zettlemoyer, and Christoph Feichtenhofer.
\newblock Demystifying clip data.
\newblock {\em International Conference on Learning Representations}, 2024.

\bibitem{yang2013saliency}
Chuan Yang, Lihe Zhang, Huchuan Lu, Xiang Ruan, and Ming-Hsuan Yang.
\newblock Saliency detection via graph-based manifold ranking.
\newblock In {\em Proceedings of the IEEE conference on Computer Vision and Pattern Recognition}, pages 3166--3173, 2013.

\bibitem{encoder}
Chao Yi, Lu Ren, De-Chuan Zhan, and Han-Jia Ye.
\newblock Leveraging cross-modal neighbor representation for improved clip classification.
\newblock In {\em Proceedings of the IEEE/CVF Conference on Computer Vision and Pattern Recognition}, pages 27402--27411, 2024.

\bibitem{cliploramaiu}
Maxime Zanella and Ismail Ben~Ayed.
\newblock Low-rank few-shot adaptation of vision-language models.
\newblock In {\em Proceedings of the IEEE/CVF Conference on Computer Vision and Pattern Recognition}, pages 1593--1603, 2024.

\bibitem{zhang2022tip}
Renrui Zhang, Wei Zhang, Rongyao Fang, Peng Gao, Kunchang Li, Jifeng Dai, Yu Qiao, and Hongsheng Li.
\newblock Tip-adapter: Training-free adaption of clip for few-shot classification.
\newblock In {\em European conference on computer vision}, pages 493--510. Springer, 2022.

\bibitem{zheng2024dreamlip}
Kecheng Zheng, Yifei Zhang, Wei Wu, Fan Lu, Shuailei Ma, Xin Jin, Wei Chen, and Yujun Shen.
\newblock Dreamlip: Language-image pre-training with long captions.
\newblock In {\em European Conference on Computer Vision}, pages 73--90. Springer, 2024.

\bibitem{zhou2022extract}
Chong Zhou, Chen~Change Loy, and Bo Dai.
\newblock Extract free dense labels from clip.
\newblock In {\em European Conference on Computer Vision}, pages 696--712. Springer, 2022.

\bibitem{coop}
Kaiyang Zhou, Jingkang Yang, Chen~Change Loy, and Ziwei Liu.
\newblock Learning to prompt for vision-language models.
\newblock {\em International Journal of Computer Vision}, 130(9):2337 -- 2348, 2021.

\bibitem{zhu2014saliency}
Wangjiang Zhu, Shuang Liang, Yichen Wei, and Jian Sun.
\newblock Saliency optimization from robust background detection.
\newblock In {\em Proceedings of the IEEE conference on Computer Vision and Pattern Recognition}, pages 2814--2821, 2014.

\end{thebibliography}
}

\end{document}